\documentclass[11pt,a4paper]{article}

\usepackage[T1]{fontenc}
\usepackage[utf8]{inputenc}
\usepackage{CJKutf8}
\usepackage{times}

\usepackage[margin=2.4cm,headheight=38pt,headsep=18pt,footskip=28pt]{geometry}
\usepackage{setspace}
\usepackage[table,dvipsnames]{xcolor}
\usepackage{colortbl}
\usepackage{graphicx}
\usepackage{booktabs}
\usepackage{array}
\usepackage{multirow}
\usepackage{tabularx}

\usepackage{amsmath,amssymb}

\definecolor{AMapBlue}{HTML}{48B8F0}
\definecolor{AMapDeepBlue}{HTML}{0877C9}
\definecolor{AMapSky}{HTML}{EAF8FF}
\definecolor{AMapLine}{HTML}{9EDBFA}
\definecolor{TextBlack}{HTML}{20242A}

\usepackage[numbers,sort&compress]{natbib}

\usepackage{enumitem}
\setlist{nosep,leftmargin=1.5em}
\usepackage{caption}
\usepackage{fancyhdr}
\usepackage{lastpage}
\usepackage{titlesec}
\usepackage[most]{tcolorbox}
\usepackage{mdframed}
\usepackage[colorlinks=true,linkcolor=AMapDeepBlue,citecolor=AMapDeepBlue,urlcolor=AMapDeepBlue]{hyperref}

\newcommand{\reportname}{ABot-World-0}
\newcommand{\reportlogo}{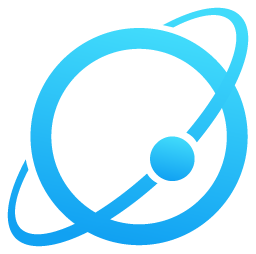}
\newcommand{\reportlogoheight}{18pt}

\newcommand{\orgname}{AMAP CV Lab \textbar{} Alibaba Group}
\newcommand{\reportcenter}[1]{\raisebox{\dimexpr(\depth-\height)/2\relax}{#1}}
\newlength{\reporttextbaselineoffset}
\newcommand{\reporttextalign}[1]{\raisebox{\reporttextbaselineoffset}{#1}}

\renewcommand{\headrulewidth}{1.2pt}
\renewcommand{\footrulewidth}{0.8pt}
\renewcommand{\headrule}{\hbox to\headwidth{\color{AMapBlue}\leaders\hrule height \headrulewidth\hfill}}
\renewcommand{\footrule}{\hbox to\headwidth{\color{AMapLine}\leaders\hrule height \footrulewidth\hfill}}

\titleformat{\section}
  {\Large\bfseries\color{AMapDeepBlue}}
  {\thesection}{0.75em}{}
  [\vspace{0.32em}{\color{AMapLine}\titlerule[0.6pt]}]
\titleformat{\subsection}
  {\large\bfseries\color{AMapDeepBlue}}
  {\thesubsection}{0.65em}{}
\titleformat{\subsubsection}
  {\normalsize\bfseries\color{AMapDeepBlue}}
  {\thesubsubsection}{0.55em}{}
\titlespacing*{\section}{0pt}{1.55em plus 0.25em minus 0.1em}{0.95em}
\titlespacing*{\subsection}{0pt}{1.25em plus 0.2em minus 0.1em}{0.55em}
\titlespacing*{\subsubsection}{0pt}{1.05em plus 0.15em minus 0.08em}{0.45em}

\newtcolorbox{abstractbox}{
  enhanced,
  breakable,
  colback=AMapSky,
  colframe=AMapBlue,
  coltitle=white,
  colbacktitle=AMapBlue,
  title=Abstract,
  fonttitle=\Large\bfseries,
  halign title=center,
  toptitle=5pt,
  bottomtitle=5pt,
  fontupper=\large,
  boxrule=0.8pt,
  arc=1.5mm,
  left=12pt,
  right=12pt,
  top=10pt,
  bottom=10pt,
  before skip=1.2em,
  after skip=1.4em
}

\newmdenv[
  backgroundcolor=gray!8,
  linewidth=0pt,
  innerleftmargin=10pt,
  innerrightmargin=10pt,
  innertopmargin=5pt,
  innerbottommargin=5pt,
  skipabove=5pt,
  skipbelow=5pt
]{cnblock}

\definecolor{DimAction}{HTML}{0877C9}
\definecolor{DimVisual}{HTML}{2E9E5B}
\definecolor{DimPhysics}{HTML}{E07B00}
\definecolor{DimMemory}{HTML}{8E44AD}

\begin{document}
\thispagestyle{fancy}

\begin{center}
  {\bfseries\color{AMapDeepBlue}\LARGE Infinite Interactive World Rollout on a Single Desktop GPU}

  \vspace{0.2em}
  {\large ABot-World Team}

  \vspace{-0.1em}
  {\small\color{TextBlack!65}July 2026}
\end{center}

\vspace{-1em}

\newcommand{\lead}{\textsuperscript{\ensuremath{\dagger}}}
\newcommand{\leadfootnotemark}{%
  \begingroup
  \hypersetup{linkcolor=TextBlack}%
  \renewcommand{\thefootnote}{\ensuremath{\dagger}}%
  \footnotemark
  \endgroup
}
\newcommand{\team}[2]{%
  \vspace{0.4em}
  \noindent
  \textbf{#1:} #2\par
}
\begin{abstractbox}

We present \textit{ABot-World-0}, an action-conditioned video world model for real-time, long-horizon closed-loop interaction, supported by a multi-source data infrastructure spanning AAA games, simulation engines, and internet videos to learn controllable world dynamics.
\textit{WorldExplorer} performs agent-driven collection guided by training feedback, while a unified pipeline applies 14 deterministic quality checks, VLM-based assessment, and synchronized action and text annotation.
We progressively distill a bidirectional action-conditioned teacher into a causal student through teacher forcing and ODE distillation, and introduce \textit{LongForcing} to align long student self-rollouts with an extended-horizon teacher, mitigating accumulated distribution shift and autoregressive drift.
Raw keyboard actions provide a unified control interface for scene roaming and third-person character interaction, while reference-character memory provides persistent appearance cues for identity consistency during third-person rollouts.
For deployment, we co-design a streaming inference stack with a lightweight VAE decoder, efficient attention, memory-aware scheduling, and low-bit DiT inference.
Across optimized low-bit configurations, \textit{ABot-World-0} streams 720P video at up to 16 FPS on a single NVIDIA RTX 5090 desktop GPU, with 1.2\,s action-to-first-frame latency and approximately 19\,GiB peak VRAM.
Experiments on WorldRoamBench and extended interactive rollouts demonstrate competitive controllability and coherent long-horizon world evolution.
\href{https://github.com/amap-cvlab/ABot-World}{https://github.com/amap-cvlab/ABot-World}

\end{abstractbox}

\begin{figure}[htbp] 
    \centering 
    \includegraphics[width=1\textwidth]{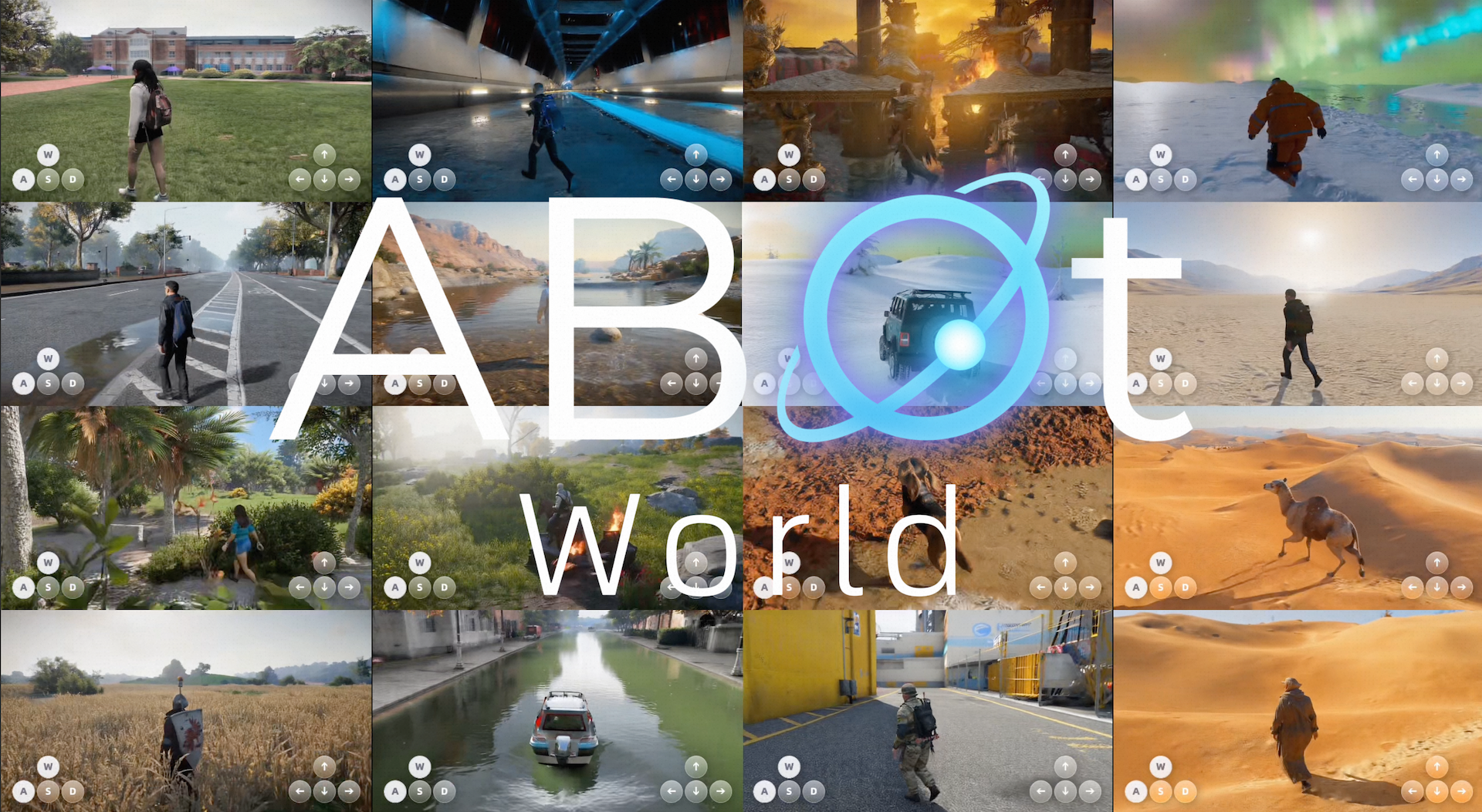}
    \caption{\textit{ABot-World-0} turns a single NVIDIA RTX 5090 GPU into a real-time interactive world simulator, enabling infinite action-conditioned world rollout at 720P and up to 16\,FPS with 1.2\,s action-to-first-frame latency within a peak-VRAM budget of approximately 19\,GiB.} 
    \label{fig:abot-world-0} 
\end{figure}

\section{Introduction}

Artificial intelligence is moving from generating static content toward generating worlds that can be entered, controlled, and continuously evolved.
The goal is not a sequence of visually plausible clips, but a persistent generative environment in which actions change state, the resulting observations inform subsequent actions, and the world remains coherent as interaction continues.
World models~\cite{kim2020gamegan,bruce2024genie,google2025genie3,valevski2025gamengen,decart2024oasis,nvidia2025cosmos,hong2025relic} are therefore a step toward general-purpose simulators: they must represent the current state of a world, predict how it evolves under interventions, and provide a substrate for planning, learning, creation, and embodied intelligence.
This is a systems-level challenge rather than a single generation objective.
Visual fidelity remains necessary, but controllability, state persistence, rollout stability, latency, throughput, and memory footprint must be optimized jointly.
Rapid advances in large-scale video generation~\cite{ho2022video,singer2022makeavideo,hong2022cogvideo,blattmann2023align,villegas2023phenaki,kondratyuk2024videopoet,kong2024hunyuanvideo,polyak2024moviegen,wan2025wan} provide increasingly powerful visual and dynamical priors, yet turning those priors into responsive, accessible worlds requires solving the closed loop.

Recent systems have established important parts of this agenda.
Genie~\cite{bruce2024genie} showed that playable environments can be learned from internet gameplay through inferred latent actions, while Genie 2 and Genie 3~\cite{parkerholder2024genie2,google2025genie3} demonstrated increasingly capable real-time dynamic environments.
GameNGen~\cite{valevski2025gamengen}, Oasis~\cite{decart2024oasis}, WHAM~\cite{kanervisto2025wham}, Runway GWM~\cite{runway2025gwm1}, Cosmos~\cite{nvidia2025cosmos}, and Waymo World Model~\cite{jiang2026waymoworld} explore complementary points in the design space of games, open worlds, physical AI, and driving.
Yet scaling a video model alone does not resolve the four coupled bottlenecks of interactive world modeling: obtaining broad, temporally coherent data with reliable action supervision; representing user intent across both camera navigation and embodied character control; preventing the generated history from drifting as it becomes the next input; and deploying the entire generation-and-decoding stack at interactive speed on practical hardware.
A system can be strong on any one axis while remaining unusable as a local, persistent world simulator.

The data problem is especially fundamental.
Passive internet video offers visual diversity but rarely exposes synchronized controls; game recordings provide exact inputs but can be stylistically narrow; simulations offer geometry and controllability but require deliberate trajectory design.
We treat these sources as complementary rather than interchangeable.
Our data infrastructure combines AAA game data, simulation-engine data, and real-world internet video.
\textit{WorldExplorer}, an agent-driven collection system, produces synchronized multimodal game and simulation trajectories and reallocates collection effort from training feedback.
A source-aware but unified processing pipeline then performs 14 deterministic checks across six quality dimensions, VLM-based semantic assessment, action annotation, and structured language annotation.
This turns dataset construction into a closed-loop part of world-model development: the data distribution is not merely collected once, but is actively shaped to expose the motion, viewpoint, environment, and action regimes in which the model remains weak.

We build on this infrastructure with \textit{ABot-World-0}, an action-conditioned video world model designed as an end-to-end answer to the control--consistency--efficiency coupling.
The model uses a unified, frame-synchronous keyboard action interface rather than a separate latent-action interface; source-native controls and pose-derived pseudo-actions are mapped into the same action space during data construction, which is naturally available to users at inference time.
A unified action representation covers observer-style scene roaming and actor-style character motion, while reference-character memory provides persistent appearance information for long third-person rollouts.
These choices make the control channel part of the dynamics model itself rather than an external post-processing interface.

The learning pipeline separates visual-dynamics quality from the causal constraint required online.
We first train a bidirectional teacher on the multi-source action-video corpus, where full-horizon generation provides a high-quality target for action-conditioned dynamics.
We then progressively convert it into a causal student through teacher forcing and ODE distillation, which preserves the deployable information pattern while reducing the denoising budget.
Crucially, local distillation does not by itself solve closed-loop generation: every student prediction changes the visual context for subsequent predictions, causing a growing mismatch between short-horizon training states and long-horizon inference states.
We introduce \textit{LongForcing}, a final distribution-matching stage that supervises long student self-rollouts with an extended-horizon bidirectional teacher.
By extending distribution-level supervision over student self-rollouts, \textit{LongForcing} provides corrective signals for long-horizon rollout contexts that short-horizon objectives cover only weakly.

Real-time interaction also requires systems co-design beyond few-step sampling.
\textit{ABot-World-0} couples its causal model with a lightweight VAE decoder, memory-aware module scheduling, low-bit DiT inference, efficient low-precision attention, and bounded local-context KV caching.
Across its optimized low-bit operating envelope, the resulting \textbf{720P} streaming pipeline reaches up to \textbf{16 FPS} with \textbf{1.2\,s} action-to-first-frame latency on a single NVIDIA RTX 5090 while operating within a peak-VRAM budget of approximately \textbf{19\,GiB}.
The reported latency covers the complete generation-and-decoding path from receiving a user action to making the first decoded response frame available, rather than reporting sampling speed in isolation.

Our contributions are fourfold:
\begin{itemize}
    \item We present \textit{ABot-World-0}, a unified action-conditioned video world model that uses raw keyboard inputs to support both scene roaming and character control, with reference-character memory providing complementary appearance cues during third-person interaction.
    \item We develop a multi-source data infrastructure for interactive world modeling, including \textit{WorldExplorer}'s training-feedback-driven collection loop, synchronized multimodal capture, multi-stage quality control, and unified action and text annotation.
    \item We introduce a progressive bidirectional-to-causal training pipeline and \textit{LongForcing}, which extends distribution-level teacher supervision to long student self-rollouts to mitigate accumulated autoregressive drift.
    \item We demonstrate a deployment-oriented streaming stack whose optimized low-bit operating envelope delivers 720P output at up to 16 FPS with 1.2\,s action-to-first-frame latency within a peak-VRAM budget of approximately 19\,GiB on one RTX 5090, and evaluate controllability, visual quality, physical plausibility, and temporal memory on WorldRoamBench together with extended interactive rollouts.
\end{itemize}

Together, these components frame interactive world modeling as a full-stack capability rather than a single generative objective.
\textit{ABot-World-0} is a step toward local visual simulators that can be continuously explored, controlled, and improved for interactive creation, agent learning, and embodied-AI research.

\section{Related Work}

\subsection{Bidirectional Video Generation}
\label{ssec:bidir}

Recent advances in generative modeling have greatly improved the quality of video synthesis, with diffusion models becoming a dominant paradigm~\cite{ho2022video,singer2022makeavideo,blattmann2023align}.
Early video diffusion models commonly adopt U-Net-based architectures, while recent methods increasingly employ diffusion transformers with spatiotemporal attention mechanisms~\cite{lu2024vdt,ma2024latte,kong2024hunyuanvideo,polyak2024moviegen,ma2025step,yang2024cogvideox}.
By modeling spatial and temporal dependencies over the whole clip, these methods~\cite{kong2024hunyuanvideo,wan2025wan,chen2024videocrafter2} achieve strong performance in visual quality and temporal coherence.
Such full-clip bidirectional modeling is also well suited to tasks such as interpolation, infilling, and bounded generation, where multiple temporal constraints need to be considered.

However, such full-clip modeling is often enabled by bidirectional attention, where each frame can access both past and future context.
While beneficial for temporal consistency, this design couples frames across the entire sequence, making it difficult to generate and output frames incrementally in low-latency or streaming scenarios.
This limitation motivates recent causal or autoregressive video diffusion methods, which attempt to convert or distill bidirectional video diffusion models into sequential generators, trading part of the global context for more efficient online generation.

\subsection{Autoregressive Video Generation}
\label{ssec:autoregressive}

To overcome the limitations of non-causal models, autoregressive (AR) video generation methods~\cite{wu2021godiva,hong2022cogvideo,ge2022long,kondratyuk2024videopoet,yan2021videogpt} produce frames sequentially, conditioning each new segment only on previously generated content.
By enforcing a causal constraint, these methods enable low-latency streaming and, when paired with bounded context or caching, reduce the memory overhead associated with long-video synthesis.
Early AR models primarily relied on discrete token-based transformers, while recent works~\cite{chen2024diffusionforcing,jin2024pyramidflow,yin2025causvid,huang2025selfforcing,liu2025rolling} have integrated causal structures into diffusion and flow-based frameworks~\cite{lipman2022flow,ho2022video,liu2022flow} to achieve superior visual quality.

The causal structure of AR generation makes it well suited for practical long-video synthesis, as it enables streaming generation.
This makes AR generation an appealing paradigm for scalable video generation.
In this work, we therefore focus on the AR paradigm and address the key challenge of improving long-horizon stability.

\subsection{Long Video Generation}
Autoregressive models suffer from drift beyond the training horizon, manifesting as progressive quality degradation, identity inconsistency, motion stagnation, or collapse to static frames after a modest number of autoregressive rollouts.

Temporal drift is closely tied to exposure bias and error accumulation in AR rollouts.
To mitigate this, methods like Diffusion Forcing~\cite{chen2024diffusionforcing}, PA-VDM~\cite{xie2025progressive}, and Rolling Forcing~\cite{liu2025rolling} employ heterogeneous noise scheduling during training~\cite{chen2025skyreelsv2, sun2025ardiffusion, kodaira2026streamdit}, while training-free approaches like FIFO-Diffusion~\cite{kim2024fifo} extend pretrained models via sliding latent windows but still suffer eventual degradation without explicit rollout alignment.
Alternatively, Self-Forcing and its variants like LongLive~\cite{yang2025longlive}, Self-Forcing{++}~\cite{cui2025selfforcingpp} supervise models under their own rollout distribution using causal attention, KV re-caching, and distillation for long video generation.
Subsequent work further addresses fine-grained mismatches: Causal Forcing~\cite{zhu2026causalforcing} tackles architectural gaps between bidirectional teachers and causal students, Context Forcing~\cite{chen2026contextforcing} leverages long-context teachers with Slow--Fast Memory, HiAR~\cite{zou2026hiar} performs hierarchical denoising at matched noise levels, and Diagonal Distillation~\cite{liu2026diagdistill} jointly optimizes temporal chunks and denoising steps.
In this work, we use a progressive bidirectional-to-causal pipeline: a teacher trained on long action-conditioned sequences provides high-quality dynamics supervision, and \textit{LongForcing} extends that supervision to long student self-rollouts to reduce accumulated drift.

Even with rollout-aligned objectives, long-range drift persists because dense attention over full history is infeasible.
While naive truncation of sliding windows or KV caches often causes identity loss and motion stagnation, recent memory-centric methods improve stability via RoPE stabilization, deep attention sinks, structured KV memory~\cite{cui2026lol, yi2025deepforcing, zhao2026relax, kim2026memrope}.
LongLive-2.0~\cite{chen2026longlive} further preserves identity across prompt switches using multi-shot attention sinks, while training-free extensions (e.g., FLEX~\cite{li2026train}, PackForcing~\cite{mao2026packforcing}) alleviate drift through positional correction or cache partitioning.
Consequently, long video generation has evolved from merely causalizing diffusion models into a comprehensive challenge spanning rollout alignment, memory design, positional extrapolation, distillation, and efficient deployment.

\subsection{Interactive Video Generation}

Interactive video generation conditions video dynamics on user actions, providing a foundation for interactive worlds, game-like simulators, and embodied AI benchmarks~\cite{parkerholder2024genie2, feng2026matrix, he2025matrix, hong2025relic}.
A common approach encodes 6-DoF camera extrinsics with dedicated networks~\cite{wang2024motionctrl, bai2025recammaster, huang2025spacetimepilot}, while another injects dense Pl\"{u}cker ray maps or frustum features aligned with latent tokens~\cite{van2024generative, he2024cameractrl, zhang2025dualcamctrl, zhou2025stable}.
These methods typically represent camera motion using calibrated trajectories in a global or initial-frame coordinate system.
Over long rollouts, accumulated poses may move beyond the training distribution, while periodic re-anchoring can introduce inconsistencies across temporal segments.
Camera-aware positional-encoding methods~\cite{li2026cameras, li2026rerope, wang2026bullettime} incorporate camera geometry or pose into token positional representations.
In contrast, we condition generation on discrete keyboard actions that specify local, incremental controls for both character movement and camera motion, rather than calibrated global camera trajectories.
These actions are interpreted relative to the current character or camera state and drawn from a fixed action space, keeping the control representation bounded and naturally aligned with real-time user interaction during long-horizon rollouts~\cite{feng2026matrix, he2025matrix, hong2025relic}.

\subsection{Real-Time Video Generation}

Real-time interactive video generation is governed by two coupled metrics: \emph{latency}, the delay between a user action and the corresponding visual update, and \emph{throughput}, the sustained frame rate the system can maintain over long rollouts~\cite{hong2025relic, lin2026autoregressive, yang2025longlive, hacohen2024ltx}.
Latency is shaped primarily by the denoising budget and the temporal unit of generation, as standard diffusion requires many score-network passes per frame and distilled few-step models~\cite{yin2024dmd, yin2024improved, song2023consistency, yin2025causvid, huang2025selfforcing, lin2026autoregressive, yang2025longlive, chen2026longlive} reduce this cost.
Frame-wise autoregressive designs~\cite{yin2025causvid, valevski2025gamengen, yang2025longlive} minimize action-to-pixel latency because each new control can influence the very next emitted frame.
However, they require one full denoising cycle per frame, which sharply limits throughput and increases sequential dependence across the rollout.
Chunk-wise generation~\cite{teng2025magi, kodaira2026streamdit, shin2025motionstream} instead denoises a chunk of frames jointly, amortizing attention and VAE decoding to achieve much higher FPS and stronger short-range temporal coherence, at the cost of a bounded control delay within the current chunk.
Throughput, in turn, depends on how efficiently each temporal unit is produced and decoded: amortizing attention and VAE decoding across multiple frames within each chunk raises FPS, while low-bit weights and KV caches~\cite{xi2026quant,chen2026longlive}, kernel fusion, and sequence or tensor parallelism~\cite{hong2025relic,jacobs2023deepspeed,yang2026swiftfusion} reduce memory and execution overhead.
Asynchronous streaming decode~\cite{chen2026longlive} can additionally overlap generation with display.
In this work, we advocate for a deployment-oriented formulation of interactive video generation that jointly optimizes latency, throughput, and memory efficiency.
Our method adopts a chunk-wise streaming pipeline with progressive visual delivery.
By co-designing lightweight VAE decoding, memory-aware scheduling, low-bit DiT inference, efficient attention, and optimized positional encoding, our optimized deployment envelope supports 720P interactive streaming at up to $16\,\mathrm{FPS}$ with $1.2\,\mathrm{s}$ action-to-first-frame latency within a peak-VRAM budget of approximately $19\,\mathrm{GiB}$ on a single NVIDIA RTX 5090 desktop GPU.


\section{Data Infrastructure}\label{sec:data_infrastructure}

\begin{figure}[htbp] 
    \centering 
    \includegraphics[width=0.95\textwidth]{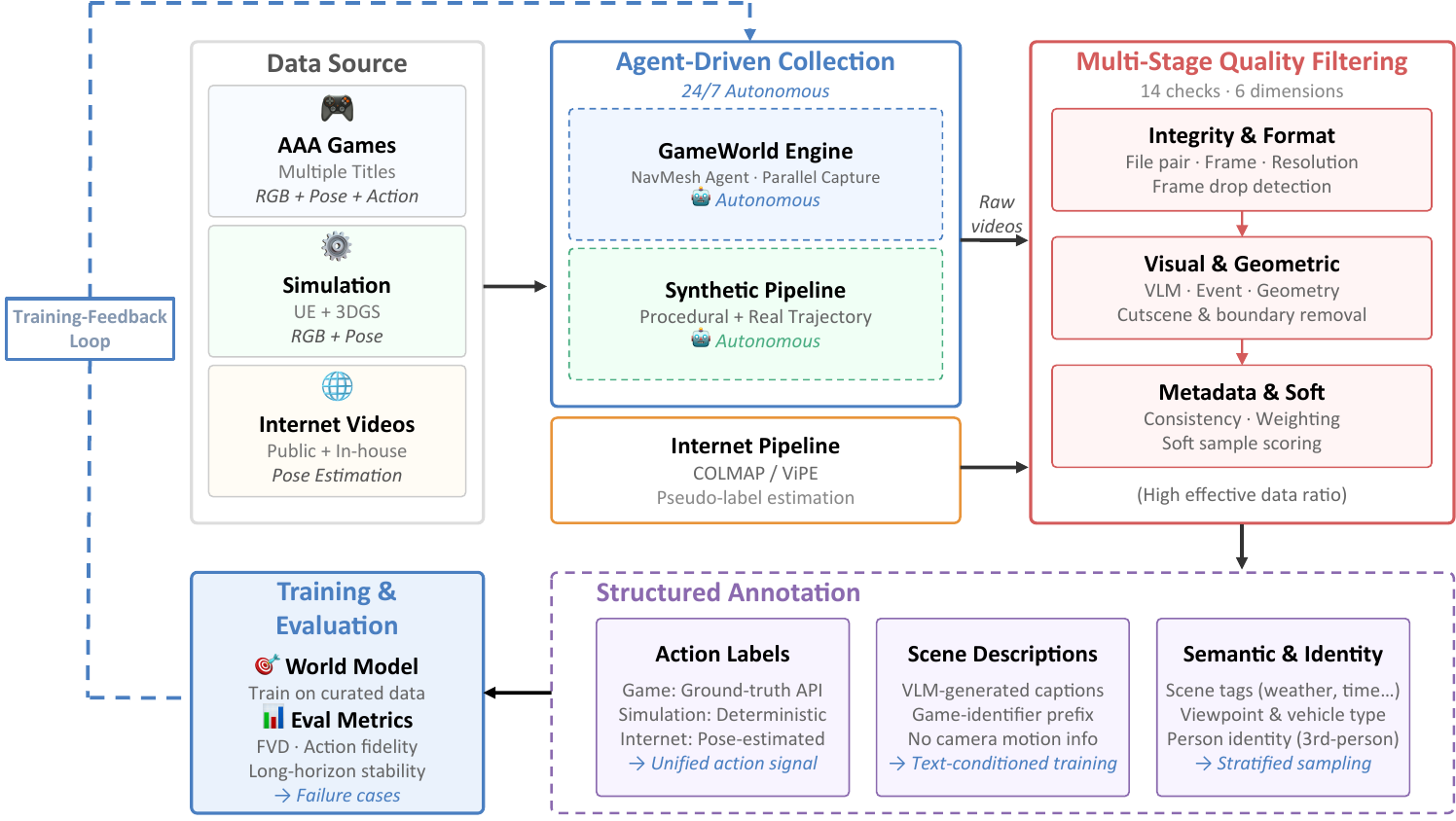}
    \caption{\textbf{Data Pipeline Overview.}
    An end-to-end pipeline couples agent-driven and internet-scale collection, multi-stage quality filtering, and structured multimodal annotation.
    Curated data trains the world model, whose evaluation feeds a training-feedback loop that drives targeted re-collection and continual data refinement.} 
    \label{fig:datapipe} 
\end{figure}

\subsection{Overview}\label{sec:overview}

\textit{ABot-World} treats data production as an integral part of model development rather than as a one-time preprocessing stage.
The construction of a high-fidelity, action-controllable world model fundamentally depends on large-scale, high-quality training data spanning diverse environments, viewpoints, and motion patterns.
Our training corpus draws from three complementary sources: AAA game data, simulation engine data, and real-world internet data.
Detailed characteristics and collection methods for each source are described in Section~\ref{sec:data_collection}.

As shown in Figure~\ref{fig:datapipe}, our data infrastructure processes all three sources through three progressive stages:

\textbf{Collection.}
Game and simulation data are collected via \textit{WorldExplorer}, our unified agent-driven system (Section~\ref{sec:worldexplorer_data_closed_loop_collection_system}), which navigates virtual environments and captures synchronized multimodal signals with closed-loop training feedback.
Internet data undergoes a separate video segmentation and pose estimation pipeline (Section~\ref{sec:internet_data_collection}).

\textbf{Quality Filtering.}
Raw collected data passes through a multi-stage filtering pipeline (Section~\ref{sec:quality_filtering}) that combines 14 deterministic signal checks spanning six quality dimensions with vision-language model (VLM)-based semantic assessment, eliminating technically flawed and semantically invalid samples.

\textbf{Annotation.}
Surviving clips are enriched with action labels, structured natural language scene descriptions, and semantic tags (Section~\ref{sec:data_annotation}) for both action-conditioned and text-conditioned training.
The pipeline follows a unified, source-independent acquisition and processing design while preserving source-native raw control signals when available and converting source-specific action information into a canonical representation for model training.

Our data infrastructure is distinguished from existing approaches by three key advantages:

\begin{itemize}
  \item \textbf{Fully automated agent-driven collection} with training-feedback-driven adaptive rebalancing, transforming data production from passive batch processing into an active servo system tightly coupled to training objectives.

  \item A \textbf{comprehensive quality control pipeline}---14 distinct checks spanning six quality dimensions supplemented by VLM-based semantic assessment---is applied systematically to maximize the effective data ratio.

  \item The combination of deterministic action annotation (for game and simulation data via direct API access and trajectory design) with a \textbf{unified multi-source processing flow} provides high-quality labels, while real-world videos contribute diversity through pose-estimated pseudo-labels.
\end{itemize}

\subsection{Data Collection}\label{sec:data_collection}

\subsubsection{Data Sources and Collection Paradigms}\label{sec:data_sources_and_collection_paradigms}

Our training data spans three complementary sources.
AAA game data provides rich, high-fidelity 3D environments with precisely synchronized action and observation signals, though inherently limited to specific titles and visual styles.
Simulation engine data offers geometrically precise synthetic data with deterministic action labels and full environmental control; notably, our street aerial photography and street-scanning data reconstructed via \textit{ABot-3DGS}~\cite{amap2023yunjing,sun2026abot} yield photorealistic real-world scenes derived from proprietary, non-public assets.
Real-world internet data introduces natural camera dynamics, diverse lighting, and domain generalization signals difficult to replicate synthetically, but lacks ground-truth action labels and requires noisy pose estimation.

Three collection paradigms have been developed for interactive environment data collection, each with distinct trade-offs.
Manual recording (human gameplay) produces authentic, naturalistic behavior but suffers from severe scalability bottlenecks, fragmented trajectories, and high labor cost.
Rule-based automated collection (scripted exploration in both game and simulation environments) operates continuously with deterministic annotations at low marginal cost, but is fundamentally constrained by its pre-defined scripts---it cannot autonomously discover novel scenarios, adapt its strategy based on training feedback, or dynamically rebalance data distribution.
Agent-driven collection, where reinforcement-learning or imitation-learning agents autonomously navigate interactive environments, enables purposeful, goal-directed exploration and produces smooth continuous trajectories at scale, while remaining prone to narrow exploration or reward-hacking failure modes and fundamentally inapplicable to passively consumed data such as internet videos.

\subsubsection{WorldExplorer: Data Closed-Loop Collection System}\label{sec:worldexplorer_data_closed_loop_collection_system}

To address the limitations of existing paradigms---particularly the rigidity of rule-based scripts and the narrow exploration and lack of training feedback in conventional agent-driven approaches---we develop \textit{WorldExplorer}, a unified agent-driven collection system applicable to both game and simulation environments.
\textit{WorldExplorer} provides a modular, environment-agnostic architecture with source-specific adapters for each backend, comprising four core components.

\textbf{Navigation Agent.}
A navigation agent autonomously traverses virtual environments through a multi-phase goal selection strategy, progressively relaxing exploration criteria---from prioritizing entirely unexplored regions, to searching nearby areas, to a forward-movement fallback with collision detection---achieving high scene coverage with low revisit rates.
This strategy applies to any interactive virtual environment with a navigable mesh, whether a live game world or a reconstructed 3DGS scene.

\textbf{Parallel Capture Pipeline.}
Video frames, camera parameters (position, rotation, field of view (FOV), focal length), control inputs, environment state, and metadata are acquired simultaneously through parallel pipelines.
All modalities are synchronized via millisecond-precision timestamps, with cross-modal alignment error below $33\,\mathrm{ms}$ at $30\,\mathrm{FPS}$, and each recording session is segmented post-hoc into training clips.

\textbf{Task Template System.}
Collection is organized into structured \textbf{task categories} to ensure behavioral distribution completeness---ranging from standard navigation and free exploration, to focused observation of landmarks, to targeted coverage of long-tail scenarios rarely occurring during natural interaction but critical for robust training.
Scene configuration is parameterized across multiple dimensions---geography, weather, time of day, traffic density, viewpoint mode, and vehicle type---enabling fine-grained control over the data distribution.

\textbf{Training-Feedback-Driven Closed-Loop.}
\textit{WorldExplorer}'s most distinguishing feature is its closed-loop, distribution-aware design.
Rather than operating with fixed collection ratios, a training monitoring module continuously tracks per-category performance metrics, a weakness diagnosis component identifies under-performing scene-action combinations via cross-dimensional scoring, and an adaptive strategy generator updates collection ratios while maintaining minimum coverage floors to prevent catastrophic forgetting.
The agent dynamically adapts through weighted task template selection and real-time scene parameter adjustment, all without manual intervention.
The full loop---\textbf{training feedback $\rightarrow$ weakness diagnosis $\rightarrow$ strategy adaptation $\rightarrow$ agent-directed production}---transforms data collection from passive batch production into an active intelligent servo system tightly coupled to training objectives throughout the model development lifecycle.

\subsubsection{Game Data Collection}\label{sec:game_data_collection}

Game data constitutes the primary and largest source of our training corpus.
\textit{WorldExplorer} is applied to multiple AAA titles across diverse genres (open-world exploration, urban driving, equestrian traversal), naturally spanning both first-person and third-person viewpoints.
Consistent with our unified, data-driven training paradigm, the pipeline supports both first- and third-person viewpoints within a shared collection and processing framework, while retaining viewpoint-specific annotations where applicable.
This design enables a single model architecture to learn action-conditioned dynamics across both viewpoints without requiring separate viewpoint-specific models.
Each recording session produces RGB video at 1920$\times$1080 resolution alongside synchronized per-frame camera poses, control signals, and environment metadata.

A distinctive advantage of game data is the viewpoint diversity it naturally provides.
First-person perspectives deliver direct ego-motion signals closely aligned with the camera's optical axis, while third-person perspectives offer observable character dynamics and surrounding spatial context that enrich the model's understanding of agent-environment interaction.
Moreover, the direct API access eliminates the label noise inherent in pose-estimation-based approaches, making game data the highest-quality supervision signal in our training corpus.

Source-native control signals are captured directly from the game's runtime API with ground-truth precision and synchronized with each video frame.
These source-specific controls are then translated into the canonical action representation used throughout the data pipeline.

\subsubsection{Simulation Data Collection}\label{sec:simulation_data_collection}

To augment trajectory diversity and provide geometric supervision signals beyond what game recordings can offer~\cite{chen2025deepverse}, we construct a synthetic pipeline using two complementary rendering backends: Unreal Engine (UE) for photorealistic scene rendering with full environmental control, and a 3D Gaussian Splatting (3DGS) backend powered by \textit{ABot-3DGS}~\cite{amap2023yunjing,sun2026abot}.
\textit{ABot-3DGS} reconstructs highly photorealistic scenes from multi-view imagery---optionally accepting LiDAR and pre-processed photogrammetry point clouds as geometric priors---and is applied to our proprietary street aerial photography and street-scanning data to produce large-scale outdoor urban environments and massive indoor scene coverage.
These geometrically accurate real-world scene representations are derived from proprietary, non-public assets.

\textit{WorldExplorer} executes trajectories on these 3D assets through two complementary modes: (a) \emph{procedural path generation}, employing geometric pattern synthesis with multi-point interpolation; and (b) \emph{real-world trajectory import}, mapping motion paths captured from physical devices into the 3D scene, preserving authentic human browsing behaviors such as repeated scanning and natural handheld jitter.
Each trajectory undergoes collision detection before rendering, and action labels are deterministically derived by projecting translational and rotational displacements onto camera basis vectors and binarizing them into discrete action signals, yielding frame-aligned annotations.

\subsubsection{Internet Data Collection}\label{sec:internet_data_collection}

For real-world internet data, agent-driven collection is not applicable since the content is passively consumed rather than interactively generated.
We instead employ a video-based pipeline to extract training signals from diverse internet sources, including driving footage, walking tours, and aerial videos.
Raw videos are first segmented into training clips based on scene boundaries and content coherence, ensuring each clip captures a consistent environment or activity.
6-DoF camera trajectories are then recovered as pseudo-labels via pose estimation methods---selected based on scene characteristics such as indoor/outdoor setting, static/dynamic content, and texture richness---to handle the diverse visual conditions encountered in internet videos.

Action intent labels are derived by projecting frame-to-frame translational and rotational displacements onto camera basis vectors and binarizing via thresholding, producing discrete action signals aligned with video frames.
While these pseudo-labels carry estimation noise absent from game and simulation ground truth, internet data substantially expands training diversity by introducing natural camera dynamics, real-world lighting variations, and domain generalization signals difficult to replicate synthetically.

\subsection{Quality Filtering}\label{sec:quality_filtering}

Raw collected data inevitably contains quality issues ranging from technical artifacts (frame drops, resolution mismatches) to semantic problems (user interface overlays, death sequences, geometry glitches).
We implement a unified multi-stage filtering pipeline comprising 14 distinct checks across six quality dimensions---(1) file integrity, (2) visual validity, (3) geometric consistency, (4) game state correctness, (5) action-label alignment, and (6) metadata quality---supplemented by VLM-based semantic assessment, to produce the final training-ready dataset.
The pipeline operates progressively across three stages, with each stage addressing a distinct category of quality concerns.
Checks execute sequentially---early-stage format validation enables rapid rejection before computationally intensive analyses.

\textbf{Stage 1: File Integrity.}
Fast deterministic checks---file pair existence, frame count consistency, resolution compliance, frame drop detection---eliminate fundamentally flawed recordings with high throughput.
Format-level errors trigger clip-level rejection as they render samples unusable for downstream action-conditioned model training.

\textbf{Stage 2: Visual Validity, Geometric Consistency \& Game State Correctness.}
Content-level analysis detects semantically invalid segments via VLM-based screening (user interface overlays, loading screens, popups, rendering anomalies), geometric anomaly detection (vertical displacement jumps for terrain clipping, camera-through-object analysis), game state signal processing (death sequence excision, cutscene and map-boundary removal), and action-label alignment verification.
Third-person character visibility is assessed and flagged (not rejected) for downstream weighting.

\textbf{Stage 3: Metadata Quality.}
Surviving clips receive metadata annotations---action-pose consistency scores, screen color shift flags---that serve as soft signals for training-time sample weighting and curriculum scheduling rather than hard rejection, allowing the model to still learn effectively from imperfect but informative training samples.

\subsection{Data Annotation}\label{sec:data_annotation}

Surviving clips are annotated for action-conditioned and text-conditioned world model training.
The pipeline is designed to be source-agnostic where possible, while exploiting source-specific signals wherever they are available.

\textbf{Action Labels.}
Action labels are obtained through source-specific mechanisms and converted into a shared canonical format.
For game data, source-native control signals are captured directly from the game's runtime API with ground-truth precision and synchronized with each video frame.
For simulation data, labels derive deterministically from designed trajectories (Section~\ref{sec:simulation_data_collection}).
For internet videos, labels derive from estimated poses via displacement projection and thresholding (Section~\ref{sec:internet_data_collection}).
These pseudo-labels carry estimation noise but enable a substantial expansion of training data diversity.
All three sources therefore provide a common action representation for unified model training.

\textbf{Scene Descriptions.}
Beyond action labels, we employ a large vision-language model (VLM) to generate structured natural language descriptions for each clip.
These descriptions serve as conditional inputs during text-conditioned generation training, enabling the model to respond to high-level semantic prompts (e.g., ``a sunny drive through a coastal highway'') in addition to low-level action controls.

The VLM generates descriptions that capture the overall scene composition, environmental characteristics, weather and lighting conditions, and notable dynamic events, while deliberately omitting camera movement information to decouple motion control from scene generation---a design choice that helps prevent the model from conflating textual descriptions with camera trajectory signals.
For game data, each caption is prefixed with a \textbf{game identifier token}, enabling the model to learn source-specific visual styles and rendering conventions.

\textbf{Semantic Tagging.}
Each clip is additionally annotated with a set of structured semantic tags extracted by the VLM, including scene attributes (indoor/outdoor, urban/rural, building types), weather conditions, time of day, vehicle type, and viewpoint mode.
These tags enable stratified sampling during training, ensuring balanced exposure to diverse environmental conditions, and facilitate targeted fine-tuning or evaluation on specific subsets.

\textbf{Person Identity.}
For third-person perspective data, we additionally extract structured person identity representations: from each third-person clip, we identify frames where the character is clearly visible and unoccluded, then crop person thumbnails oriented toward four cardinal viewing directions (frontal, rear, left-lateral, right-lateral) based on the relative camera-to-character angle.
From these directional thumbnails, we further synthesize a canonical frontal portrait via image-based face completion, producing a standardized identity reference.

These annotations support identity-conditioned generation in third-person scenarios and provide identity references for character-consistent video synthesis.
Together, the curated multi-source corpus supports the complete \textit{ABot-World} training pipeline, while model evaluation feeds subsequent \textit{WorldExplorer} collection and closes the data--model loop shown in Figure~\ref{fig:datapipe}.

\section{ABot-World-0}
\label{sec:method}

This section presents the complete \textit{ABot-World-0} pipeline.
We first introduce the general problem formulation (Section~\ref{sec:formulation}), which defines the interactive world modeling task, including the action-conditioned inputs and training objectives.
We then describe bidirectional teacher training (Section~\ref{sec:teacher}), which equips the pretrained video generator with action-conditioned dynamic modeling capability through large-scale interactive video data.
Next, we present causal distillation (Section~\ref{sec:distillation}), which transfers the bidirectional generation capability into a causal autoregressive model that supports efficient online interaction and long-horizon rollout.
To further improve rollout stability, we introduce \textit{LongForcing}, a long-horizon distribution-matching strategy that extends teacher supervision over longer temporal contexts and mitigates accumulated autoregressive drift.
Finally, we detail our inference acceleration pipeline (Section~\ref{sec:accel}).

In summary, this section presents three core technical components:
\begin{enumerate}[leftmargin=*]
    \item \textbf{An action-controllable video world model for unified interactive generation.}
    \textit{ABot-World-0} uses raw keyboard inputs as the sole action interface for both scene roaming and third-person character interaction, while reference-character memory provides complementary appearance conditioning to support character identity consistency during third-person interaction.
    \item \textbf{A progressive bidirectional-to-causal training pipeline for efficient and stable rollout.}
    Teacher forcing and ODE distillation progressively transfer the bidirectional teacher into a causal few-step student, while \textit{LongForcing} aligns long student self-rollouts with an extended-horizon teacher to mitigate accumulated distribution shift and autoregressive drift.
    \item \textbf{A full-stack inference co-design for real-time single-GPU deployment.}
    The deployment stack combines few-step diffusion with lightweight VAE decoding, low-bit DiT inference, efficient attention and positional encoding, and memory-aware scheduling to enable real-time interactive generation within the compute and memory constraints of a \textbf{single desktop GPU}.
\end{enumerate}

\subsection{Formulation}
\label{sec:formulation}

A world model fundamentally aims to predict the future state of an environment from past observations and agent interactions.
In our setting, the environment state is represented by video, and the model predicts future visual observations conditioned on previously observed frames, user actions, and multimodal conditions such as high-level textual instructions and reference images.
We formulate this task as an \emph{interactive world modeling} problem, which captures three key requirements: \emph{long-horizon video rollout}, \emph{action conditioning}, and \emph{temporal consistency} over extended horizons.

Formally, let $\mathbf{v}_{0:t-1}$ and $\mathbf{a}_{t:t+L-1}$ denote the available visual history and future action sequence, respectively.
Let $\mathbf{c}$ denote the multimodal conditions comprising a text prompt and reference images.
We aim to model
\begin{equation}
\label{eq:causal-objective}
    p_\theta\!\left(
        \mathbf{v}_{t:t+L-1}
        \mid
        \mathbf{v}_{0:t-1},
        \mathbf{a}_{t:t+L-1},
        \mathbf{c}
    \right),
\end{equation}
where $\mathbf{v}_{t:t+L-1}$ denotes the next video chunk of length $L$.
By iteratively predicting and appending future chunks, the model can autoregressively roll out long-horizon trajectories.

\begin{figure}[htbp] 
    \centering 
    \includegraphics[width=0.95\textwidth]{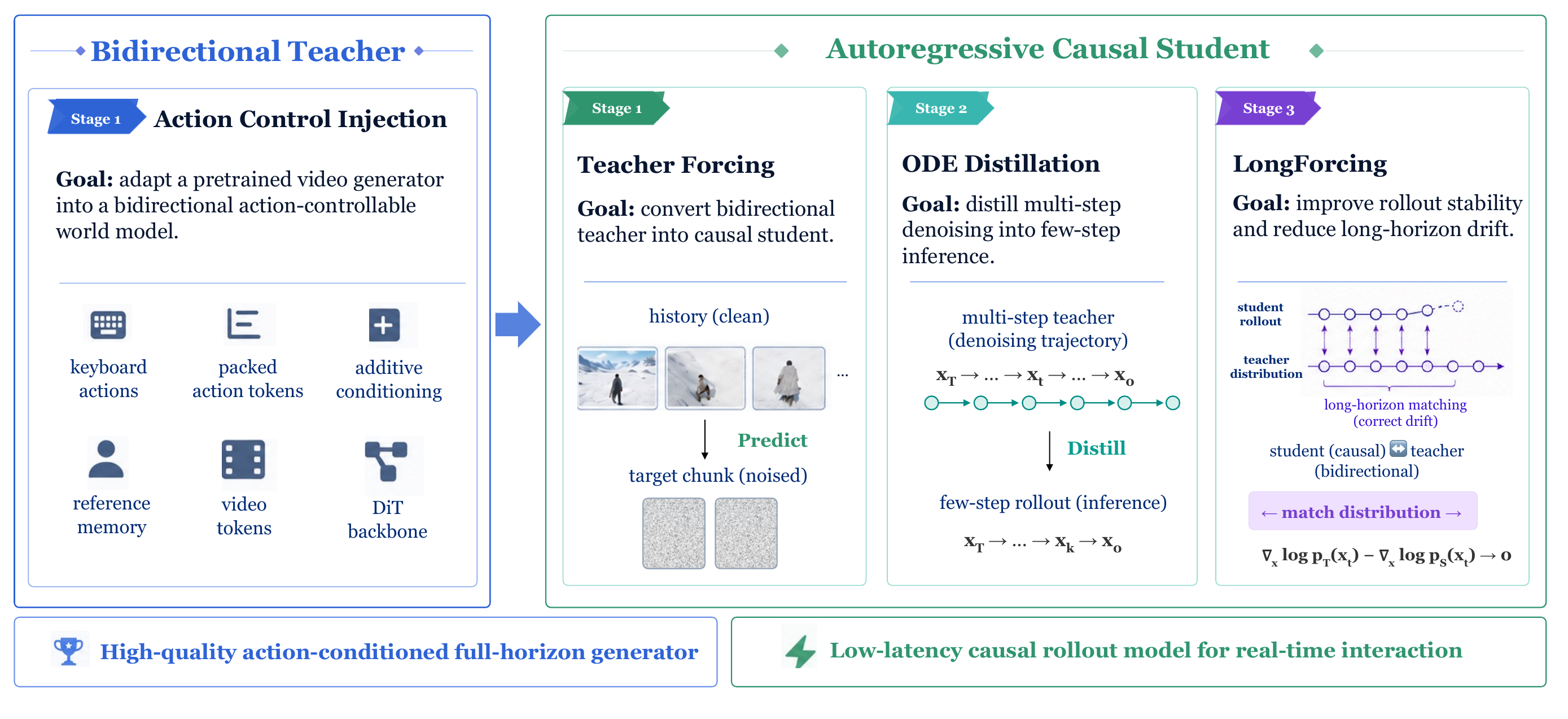}
    \caption{\textbf{Overall training pipeline of ABot-World-0.}
    We first adapt a pretrained video generator into a high-quality bidirectional action-conditioned teacher.
    The teacher is then progressively converted into a causal autoregressive student through Teacher Forcing, Causal ODE Distillation, and \textit{LongForcing}, enabling few-step low-latency inference and stable long-horizon interactive rollout.} 
    \label{fig:abot-world-pipe} 
\end{figure}

Our training pipeline consists of two main phases as shown in Figure~\ref{fig:abot-world-pipe}, each corresponding to a different conditional modeling objective:

\begin{enumerate}
    \item \textbf{Bidirectional Teacher Learning.}\quad
    In the first phase, we train a bidirectional teacher to generate a full-horizon video conditioned on an initial frame, the complete action sequence, and multimodal conditions.
    Specifically, given the initial frame $v_0$, actions $\mathbf{a}_{1:T}$, and multimodal conditions $\mathbf{c}$, the teacher is trained to model
    \begin{equation}
    \label{eq:bidirectional-objective}
        p_{\phi}^{\mathrm{bi}}\!\left(
            \mathbf{v}_{1:T}
            \mid
            v_0,
            \mathbf{a}_{1:T},
            \mathbf{c}
        \right).
    \end{equation}
    The bidirectional architecture generates the target sequence jointly, allowing information to propagate across the entire temporal horizon rather than enforcing a strictly causal dependency.
    This non-causal formulation allows full-horizon information exchange and provides a strong teacher for visual consistency, motion quality, and action alignment.
    Starting from a pretrained video generation backbone, we fine-tune the teacher on large-scale video--action data with action conditioning, reference-character memory, and long-sequence training.
    \item \textbf{Causal Student Learning via Distillation.}\quad
    In the second phase, we distill the trained bidirectional teacher into a causal autoregressive generator whose objective matches the online rollout setting in Eq.~\eqref{eq:causal-objective}.
    Specifically, the student models
    \begin{equation}
    \label{eq:student-objective}
        p_{\theta}^{\mathrm{causal}}\!\left(
            \mathbf{v}_{t:t+L-1}
            \mid
            \mathbf{v}_{0:t-1},
            \mathbf{a}_{t:t+L-1},
            \mathbf{c}
        \right),
    \end{equation}
    predicting each future video chunk using only past visual observations, the corresponding action sequence, and the multimodal conditions, without access to future frames.

    Because the bidirectional teacher relies on full-sequence generation and iterative denoising, it cannot be directly deployed for interactive autoregressive rollout.
    We therefore adopt a progressive three-stage distillation pipeline.
    First, \emph{teacher forcing} transfers the teacher's knowledge to an autoregressive student under causal visual conditioning.
    Second, \emph{ODE distillation} compresses multi-step autoregressive denoising into few-step prediction by learning the mapping from intermediate noisy latents to clean endpoints defined by the frozen Stage~1 causal model's probability-flow ODE under the same causal inputs.
    Third, we introduce \textit{LongForcing}, which performs distribution matching against an extended-horizon teacher.
    The longer supervision horizon covers later portions of student self-rollouts, where prediction errors have had more opportunities to accumulate.
    This provides distribution-level corrective supervision beyond short-horizon training and reduces long-term drift during autoregressive rollout.
\end{enumerate}

The following sections describe each stage in detail.

\subsection{Bidirectional Teacher Training with Action Control Injection}
\label{sec:teacher}

Pretrained video generation models possess rich world priors such as visual appearance, physical plausibility, and scene composition, but lack action-conditioned dynamic knowledge, \textit{i.e.}, the mapping from discrete control inputs to corresponding visual state transitions.
To bridge this gap, we train a bidirectional video generation model by injecting action control signals and reference images into the pretrained Wan2.2 backbone~\cite{wan2025wan}.
Through full-parameter fine-tuning on large-scale multi-source video--action data with temporal augmentation, our model learns the correspondence between actions and visual dynamics, providing a strong foundation for subsequent causal autoregressive adaptation.

\subsubsection{Action Control Injection}

We use raw keyboard input as the sole interactive control signal.
Specifically, each frame-level action is represented as an 8-dimensional multi-hot vector corresponding to 8 discrete keys: \texttt{W/A/S/D} for character or camera movement and \texttt{I/J/K/L} for camera rotation.
Let
\begin{equation}
    \mathbf{a}_{1:T} = \{a_1,\ldots,a_T\}, \qquad a_t \in \{0,1\}^{8},
\end{equation}
denote the per-frame action sequence for a video of length $T$.
Compared with methods that rely on estimated camera poses or learned latent actions, such raw keyboard signals are directly available from game recordings and align naturally with user intent at inference time.

To match the temporal compression of the video tokenizer, we pack every 4 consecutive frame-level actions along the channel dimension, consistent with the VAE temporal patch size of 4.
Concretely, the packed action token sequence is
\begin{equation}
    \tilde{\mathbf{a}}_{1:T/4} = \{\tilde{a}_1,\ldots,\tilde{a}_{T/4}\}, \qquad \tilde{a}_\tau \in \{0,1\}^{32},
\end{equation}
where
\begin{equation}
    \tilde{a}_\tau = \mathrm{Concat}\!\left(a_{4\tau-3}, a_{4\tau-2}, a_{4\tau-1}, a_{4\tau}\right).
\end{equation}
Thus, each temporal action token has dimension $8 \times 4 = 32$ and is aligned with one latent frame after VAE compression.

We then feed the packed action sequence into an Action Control Adapter $\mathcal{F}_{\psi}$.
We adopt a camera adapter architecture consisting of \texttt{PixelUnshuffle}, a convolution layer whose kernel size and stride are both set to the DiT spatial patch size, followed by residual convolution blocks.
The downsampling factor of \texttt{PixelUnshuffle} is chosen to match the spatial compression ratio of the VAE, so that the adapter output is strictly aligned with the spatiotemporal resolution of the DiT patchified latent tokens.
Denoting the noised video latents by $\mathbf{z}$ and their patch embeddings by $\mathrm{PatchEmbed}(\mathbf{z})$, the action-conditioned latent tokens are given by
\begin{equation}
    \hat{\mathbf{z}} = \mathrm{PatchEmbed}(\mathbf{z}) + \mathcal{F}_{\psi}(\tilde{\mathbf{a}}_{1:T/4}).
\end{equation}
This design enables direct token-wise additive injection of action information into the DiT backbone.

We use additive injection at the patchify stage throughout our model.

\subsubsection{Identity-Preserving Conditioning}

Video world models may suffer from \emph{identity drift} during long-horizon third-person rollout, where the character appearance gradually deviates despite preserving motion dynamics.
To address this issue, we introduce a reference-character memory module composed of multiple canonical reference images of the controllable character.

The reference images are encoded by the same VAE as the rollout frames and converted into identity-memory tokens, which are prepended to the video-token sequence before the DiT backbone.
We assign fixed negative temporal RoPE indices to memory tokens and non-negative indices to video tokens, separating static identity information from the generated trajectory.
Furthermore, we adopt an asymmetric memory--video attention pattern, where video tokens attend to memory tokens while memory tokens remain isolated from video tokens.
This enables persistent identity retrieval throughout autoregressive rollout and improves character consistency over long horizons.

\begin{figure}[htbp] 
    \centering 
    \includegraphics[width=0.95\textwidth]{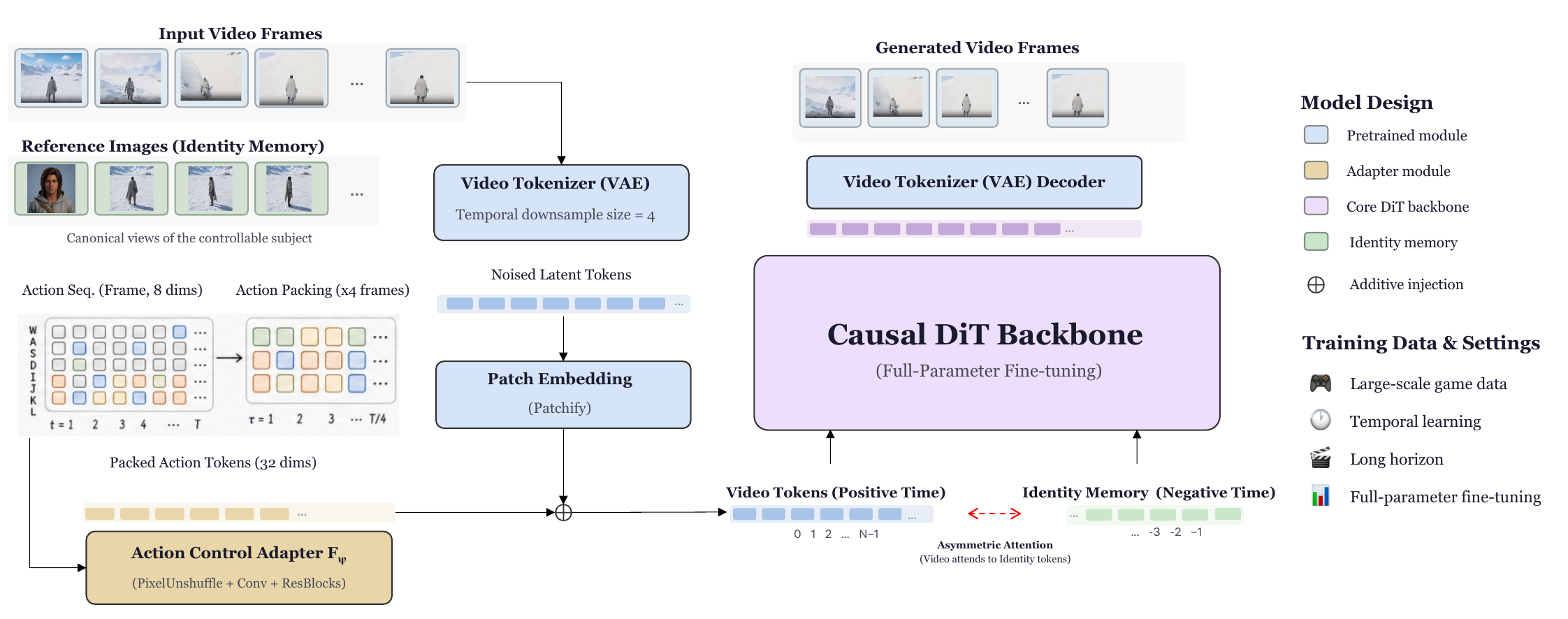}
    \caption{\textbf{ABot-World-0 Model Architecture.}
    \textit{ABot-World-0} incorporates action control and reference-character memory into a pretrained video DiT.
    Keyboard actions guide dynamics, while reference images help preserve character identity.} 
    \label{fig:abot-world-model} 
\end{figure}

\subsection{Causal Progressive Student Distillation}
\label{sec:distillation}

The bidirectional teacher cannot be directly deployed for real-time interaction due to its full-sequence generation and multi-step denoising.
We therefore distill it into a causal generator through a three-stage progressive pipeline: (1)~\textbf{Teacher forcing}, which converts the bidirectional teacher into an autoregressive student under causal visual conditioning; (2)~\textbf{ODE distillation}, which reduces the autoregressive denoising process to few-step inference while preserving generation quality; and (3)~\textbf{LongForcing}, which performs long-horizon distribution matching with an extended-horizon teacher to improve rollout stability and mitigate accumulated drift during autoregressive generation.
By exposing the student to a broader range of long-horizon student-rollout contexts where rollout errors may accumulate, \textit{LongForcing} provides corrective supervision beyond short training horizons.
Each stage builds on the previous one, progressively transforming the high-quality but non-causal teacher into an efficient causal world model suitable for real-time interactive rollout.

\subsubsection{Stage 1: Teacher Forcing}

Rather than training the causal student from scratch, we initialize it from the trained bidirectional teacher and adapt the model through a causal training objective and attention masking strategy.
While the bidirectional teacher jointly denoises the entire video sequence with access to the full temporal context, the causal student is trained to generate future video chunks using only previously observed visual context.

Specifically, during teacher forcing, the history frames are provided as clean ground-truth latents, while the target chunk is corrupted according to the diffusion process.
A causal attention mask restricts the model from accessing future visual contexts beyond the available history, thereby matching the information structure required for autoregressive inference.
This formulation retains the visual priors and long-range dynamics knowledge learned by the bidirectional teacher while adapting the model to causal generation.

Formally, for a rollout step starting at time $t$, the student models
\begin{equation}
    p_{\theta}^{\mathrm{causal}}\!\left(
        \mathbf{v}_{t:t+L-1}
        \mid
        \mathbf{v}_{0:t-1},
        \mathbf{a}_{t:t+L-1},
        \mathbf{c}
    \right),
\end{equation}
where $\mathbf{v}_{0:t-1}$ is used as clean visual conditioning and $\mathbf{v}_{t:t+L-1}$ is converted into noisy latent targets for diffusion training.
By relying on ground-truth history at this stage, the objective provides a stable adaptation from bidirectional trajectory modeling to causal autoregressive rollout.

\subsubsection{Stage 2: ODE Distillation}

The causal diffusion model obtained in the previous stage still requires iterative denoising at each autoregressive rollout step.
To reduce inference latency, we apply causal ODE distillation, which trains a few-step model to approximate the probability-flow ODE induced by the Stage~1 causal model.

A key requirement is that the reference model and the distilled model follow the same causal factorization.
We therefore freeze the Stage~1 causal diffusion model, parameterized by $\theta_{\mathrm{c}}$, and condition both models on the same causal context:
\begin{equation}
    \mathcal{C}_t
    =
    \left(
        \mathbf{v}_{0:t-1},
        \mathbf{a}_{t:t+L-1},
        \mathbf{c}
    \right),
\end{equation}
where $\mathbf{v}_{0:t-1}$ denotes the clean visual history, $\mathbf{a}_{t:t+L-1}$ denotes the target action sequence, and $\mathbf{c}$ denotes the multimodal conditions comprising a text prompt and reference images.

Let $\mathbf{z}^{\mathrm{c}}_s$ denote an intermediate latent at noise level $s$ along the probability-flow ODE trajectory of the Stage~1 causal model.
Its corresponding clean endpoint is
\begin{equation}
    \mathbf{z}^{\mathrm{c}}_0
    =
    \Phi_{\theta_{\mathrm{c}},\,s\rightarrow 0}
    \left(
        \mathbf{z}^{\mathrm{c}}_s;
        \mathcal{C}_t
    \right),
\end{equation}
where $\Phi_{\theta_{\mathrm{c}},\,s\rightarrow 0}$ denotes integration of the causal model's probability-flow ODE from noise level $s$ to the clean endpoint.
The distilled model is trained to directly predict this endpoint:
\begin{equation}
    \mathcal{L}_{\mathrm{ODE}}
    =
    \mathbb{E}_{s,\mathbf{z}^{\mathrm{c}}_s,\mathcal{C}_t}
    \left[
        \left\|
            f_{\theta}
            \left(
                \mathbf{z}^{\mathrm{c}}_s,
                s,
                \mathcal{C}_t
            \right)
            -
            \operatorname{sg}
            \left(
                \mathbf{z}^{\mathrm{c}}_0
            \right)
        \right\|_2^2
    \right],
\end{equation}
where $\operatorname{sg}(\cdot)$ denotes the stop-gradient operation.

Because the intermediate latent and its endpoint are obtained from the same causal ODE trajectory under identical conditioning, they define a consistent flow map for each autoregressive prediction step.
Moreover, the distillation target depends only on the causal context available during deployment and does not require future visual observations.
The resulting model can therefore approximate the original causal diffusion endpoint map with substantially fewer denoising steps while preserving its autoregressive rollout behavior.

\subsubsection{Stage 3: LongForcing}

After ODE distillation, the causal student can perform autoregressive rollout with only a few denoising steps.
However, it still faces a fundamental long-horizon challenge: since each prediction is conditioned on previously generated frames, small distributional errors can accumulate and gradually shift the rollout toward long-horizon student-rollout contexts that are rarely encountered during training.
These long-horizon student-rollout contexts receive limited distribution-matching supervision, causing the rollout distribution to gradually drift away from the desired world dynamics.

We therefore introduce \textit{LongForcing}, a final long-horizon distribution-matching stage that improves rollout stability by extending the temporal horizon of teacher supervision.
Unlike ODE distillation, which learns clean-endpoint flow maps under fixed causal conditioning, \textit{LongForcing} aligns the student and teacher at the level of long-horizon conditional video distributions.
The key insight is that long-horizon stability depends not only on accurate local transitions, but also on whether the closed-loop rollout distribution remains within the temporal region covered by teacher supervision.

Longer student self-rollouts expose the model to long-horizon student-rollout contexts in which small prediction errors have had more opportunities to accumulate.
\textit{LongForcing} trains on these self-generated trajectories and applies distribution-level corrective supervision from an extended-horizon teacher during the final Distribution Matching Distillation (DMD) stage~\cite{yin2024dmd}.
This encourages the student to maintain plausible long-horizon dynamics without requiring explicit frame-level trajectory matching.
Compared with short-horizon teacher supervision, the extended horizon provides distribution-level guidance over a longer temporal range and is designed to mitigate progressive degradation in visual quality and action-conditioned dynamics.

\subsection{Full-Stack Co-Design for Real-Time World Rollout}
\label{sec:accel}

Few-step distillation substantially reduces the denoising cost of each autoregressive prediction, but it is not sufficient by itself to deliver real-time interactive world rollout on a single consumer-grade GPU.
As the denoising budget is compressed, the dominant system bottlenecks shift toward chunk-wise latent decoding, transformer computation, attention and positional-encoding overhead, context-memory traffic, and runtime model residency.

We therefore formulate real-time deployment as a \emph{full-stack co-design} problem spanning the temporal generation granularity, VAE decoder, low-precision DiT execution, attention and RoPE kernels, bounded context memory, and runtime module scheduling.
Rather than optimizing an isolated operator, our objective is to jointly balance three deployment-critical properties: sustained generation throughput, action-to-first-frame latency, and peak GPU memory usage.
This co-designed inference stack enables real-time, high-resolution, long-horizon interactive generation on a single NVIDIA RTX 5090 GPU.

\subsubsection{Deployment Objective and Runtime Metrics}

\begin{table}[t]
\centering
\caption{Deployment setting and operating envelope of \textit{ABot-World-0} on a single NVIDIA RTX 5090.}
\label{tab:deployment_setting}
\begin{tabular}{@{}ll@{}}
\toprule
Item & Specification \\
\midrule
GPU & $1\times$ NVIDIA RTX 5090 \\
Batch size & $1$ \\
Resolution & $1280\times704$ \\
Inference & Chunk-wise streaming \\
Throughput & Up to $16\,\mathrm{FPS}$ \\
Action-to-first-frame latency & $1.2\,\mathrm{s}$ \\
Peak VRAM & $\leq19.3\,\mathrm{GiB}$ \\
\bottomrule
\end{tabular}
\end{table}

\textit{ABot-World-0} adopts chunk-wise causal generation, jointly producing a chunk of latent frames and decoding the complete chunk before delivering the resulting frames to the streaming runtime.
We define \emph{action-to-first-frame latency} as the wall-clock time from a user keypress until the corresponding inference chunk has been decoded and its first response frame becomes available.

The table reports the overall deployment envelope rather than a single precision-specific operating point.
Detailed speed and memory trade-offs across FP8 and more aggressive low-bit configurations are reported in Table~\ref{tab:system_breakdown}.

\subsubsection{Chunk-Wise Streaming and Runtime Co-Design}

\textbf{A lightweight VAE decoder reduces first-frame latency and peak memory usage.}
Although the DiT remains the dominant computational component, VAE decoding still contributes non-negligibly to first-frame latency and memory consumption at high resolution.
Inspired by TAEHV~\cite{boerbohan2025taehv}, we simplify and prune the decoder architecture to construct a lightweight VAE decoding path, denoted \textit{LightVAE}, substantially reducing decoding time and peak memory usage.

Because the complete latent chunk must be decoded before its response frames become available to the streaming runtime, reducing chunk decoding time directly lowers the action-to-first-frame latency.

\textbf{Memory-aware scheduling improves single-GPU deployment feasibility.}
To further reduce GPU memory pressure, we adopt memory-aware runtime scheduling inspired by the module-swapping utilities of FramePack~\cite{zhang2025framepackrepo}.
Instead of requiring all components of the inference pipeline to remain simultaneously resident on the GPU, model modules are staged according to their execution order and memory requirements.
This strategy reduces peak VRAM usage without modifying the model architecture, while limiting data-transfer overhead along the latency-critical inference path.

\textbf{Fast-RoPE reduces temporal positional-encoding overhead.}
Long-horizon autoregressive generation repeatedly applies temporal positional encoding over a moving local-attention context.
We adopt \textit{Fast-RoPE} to improve temporal positional encoding during streaming generation.
Specifically, we re-anchor temporal RoPE within the local attention window to reduce repeated RoPE computation over the full visible context as the rollout advances.
We additionally use a Triton-based RoPE kernel to accelerate the rotation operation.
Together, these optimizations reduce positional-encoding overhead during long-horizon interactive inference.

\subsubsection{Low-Precision Compute and Context-Memory Co-Design}

\textbf{Low-bit mixed-precision inference improves DiT throughput and memory efficiency.}
To reduce the computational and memory-bandwidth cost of DiT inference, we apply low-bit quantization following the deployment practices of LightX2V~\cite{lightx2v2025framework}.
FP8 serves as our default quality-oriented operating point and provides substantial improvements in GEMM throughput and memory-bandwidth efficiency.

We additionally evaluate more aggressive low-bit formats to characterize the upper-throughput region of the deployment envelope.
Concretely, the compute-intensive linear layers of the DiT backbone are quantized, while numerically sensitive components such as the VAE and text encoder remain in higher precision.
This mixed-precision design provides configurable trade-offs in inference throughput and memory usage while retaining higher precision for numerically sensitive components.

\textbf{Efficient attention kernels accelerate the dominant transformer operators.}
Attention computation constitutes a major part of the per-step latency of the diffusion transformer, particularly for high-resolution video generation with long token sequences.
We therefore adopt SageAttention2~\cite{zhang2024sageattention2} as an efficient attention backend.
It replaces conventional attention execution without requiring model retraining or architectural modification, reducing the runtime cost of a dominant transformer operator.

\textbf{Bounded KV caching prevents unbounded context growth, while cache quantization can further reduce its fixed cost.}
For long-horizon streaming generation, a naive full-history KV cache grows continuously with the generated context and eventually becomes a practical memory and bandwidth bottleneck.
\textit{ABot-World-0} avoids this unbounded growth through a bounded local-context KV cache with rolling eviction, keeping the cache footprint independent of the total rollout duration.

Nevertheless, even under a bounded-context design, the resident key and value tensors remain a substantial source of memory consumption and memory-bandwidth traffic at high resolution.
To further reduce this cost, we explore KV-cache quantization, which compresses cached key and value tensors while preserving their usability for subsequent attention computation.

Following recent advances in online vector quantization~\cite{zandieh2025turboquant}, this strategy can reduce both the fixed cache footprint and the bandwidth pressure of local-context attention.
KV-cache quantization can therefore complement low-bit DiT inference and improve the scalability of long-horizon generation under constrained GPU memory.

\subsubsection{End-to-End System Analysis}

\begin{table}[htbp]
\centering
\caption{System-level breakdown across optimization variants on a single NVIDIA RTX 5090 GPU at $1280\times704$ resolution and batch size $1$.}
\label{tab:system_breakdown}
\resizebox{\linewidth}{!}{
\begin{tabular}{lcccc}
\toprule
Configuration &
DiT time (ms/chunk) &
VAE time (ms/chunk) &
FPS $\uparrow$ &
VRAM (GiB) $\downarrow$ \\
\midrule
Base &
-- &
-- &
OOM &
OOM \\

+ SageAttention2 &
-- &
-- &
OOM &
OOM \\

+ SageAttention2 + LightVAE &
1191.081 &
78.276 &
9.117 &
20.491 \\

+ SageAttention2 + LightVAE + FP8 &
845.180 &
75.980 &
12.405 &
\textbf{15.925} \\

+ SageAttention2 + LightVAE + FP8 + Fast-RoPE &
786.871 &
\textbf{71.730} &
13.269 &
19.281 \\

+ SageAttention2 + LightVAE + MXFP6 + Fast-RoPE &
718.281 &
85.994 &
14.098 &
18.287 \\

+ SageAttention2 + LightVAE + MXFP4 + Fast-RoPE &
\textbf{638.843} &
72.957 &
\textbf{15.831} &
17.148 \\
\bottomrule
\end{tabular}
}
\end{table}

Each inference chunk contains $3$ latent frames and produces $12$ decoded video frames.
DiT and VAE times are reported per inference chunk, FPS denotes generation throughput, VRAM denotes peak GPU memory usage in GiB, and OOM denotes out of memory.
Table~\ref{tab:system_breakdown} demonstrates that real-time deployment emerges from full-stack optimization rather than the acceleration of a single operator.
Both the \textit{Base} configuration and the \textit{SageAttention2}-only variant run out of memory, showing that a faster attention kernel alone is insufficient to make the complete high-resolution pipeline deployable on a single desktop GPU.
Memory feasibility requires the joint optimization of transformer execution, video decoding, numerical precision, and runtime model residency.

Introducing \textit{LightVAE} yields the first feasible configuration, achieving 9.117 FPS with 20.491 GiB of peak VRAM.
This result indicates that the original VAE is an important bottleneck for memory feasibility, while VAE decoding remains a non-negligible component of the chunk-response latency.

Building on this configuration, quantizing the DiT backbone provides substantial gains.
FP8 reduces the DiT time from 1191.1 ms to 845.2 ms per chunk, increases generation throughput from 9.117 to 12.405 FPS, and reduces peak memory usage from 20.491 GiB to 15.925 GiB.
Adding \textit{Fast-RoPE} further reduces the DiT time to 786.9 ms and raises throughput to 13.269 FPS.
The measured peak memory increases to 19.281 GiB in this configuration, highlighting that peak VRAM is determined by the complete runtime configuration rather than by the cost of an individual operator alone.

We use FP8 as the default quality-oriented configuration, while more aggressive low-bit formats extend the upper-throughput operating envelope.
Across the optimized low-bit configurations, \textit{ABot-World-0} reaches up to 16 FPS while peak VRAM remains below 19.3 GiB.

Overall, the results validate our central systems insight: few-step generation does not automatically translate into real-time interaction.
Real-time, high-resolution world rollout requires the full-stack co-design of temporal generation, latent decoding, transformer arithmetic, attention and positional encoding, context-memory management, and runtime scheduling.
Together, these optimizations turn the distilled causal model into a practical interactive world simulator that can run continuously on a single desktop GPU.

Quantization-aware training may further improve the speed--memory--quality trade-off.
We leave this direction for future work.

\section{Evaluation}
\label{sec:eval}

We evaluate \textit{ABot-World-0} through three complementary views: quantitative comparison on WorldRoamBench, a 60-second temporal ablation of \textit{LongForcing}, and qualitative stress tests spanning hour- and day-scale rollouts, out-of-domain control, and physical interaction.
WorldRoamBench measures action controllability, trajectory following, visual quality, physical plausibility, and temporal memory, while the extended rollouts examine model behavior beyond the benchmark horizon.
The \textit{LongForcing} ablation directly evaluates visual error accumulation during autoregressive rollout, and the qualitative cases illustrate extended interaction, identity persistence, control generalization, and plausible physical effects.
Together, these experiments characterize the controllability, coherence, and state persistence of \textit{ABot-World-0} over prolonged interaction.

\subsection{WorldRoamBench Evaluation}
\label{sec:worldroambench}

To evaluate \textit{ABot-World-0} in interactive, long-horizon scenarios, we use WorldRoamBench~\cite{xu2026worldroambench}, an open-world benchmark for controllable video world models.
WorldRoamBench assesses four critical dimensions: action controllability, visual plausibility, physical consistency, and temporal memory retention.

\subsubsection{Quantitative Results}
\label{sec:worldroambench_quantitative}
Table~\ref{tab:worldroambench} reports concrete WorldRoamBench sub-dimensions only.
We compare against Genie 3~\cite{google2025genie3}, HappyOyster~\cite{alibaba2026happyoyster}, LingBot-World~\cite{robbyant2026lingbotworld}, and HY-World 1.5~\cite{tencent2025hyworld15} following the benchmark protocol.
\begin{center}
\captionsetup{type=table,hypcap=false}
\caption{WorldRoamBench quantitative comparison on selected sub-dimensions of {\color{DimAction}\textbf{Action}}, {\color{DimVisual}\textbf{Visual}}, {\color{DimPhysics}\textbf{Physics}}, and {\color{DimMemory}\textbf{Memory}}, with headers colored accordingly.
Best results are shown in \textbf{bold}, and second-best results are \underline{underlined}.}
\label{tab:worldroambench}
\small
\setlength{\tabcolsep}{4pt}
\resizebox{\linewidth}{!}{
\begin{tabular}{@{}l*{8}{c}@{}}
\toprule
\textbf{Model} & \textbf{Size} &
{\color{DimAction}\textbf{Strict Acc.}} & {\color{DimAction}\textbf{Partial Acc.}} & {\color{DimAction}\textbf{Traj. Score}} &
{\color{DimVisual}\textbf{Aesthetic}} & {\color{DimVisual}\textbf{Imaging}} &
{\color{DimPhysics}\textbf{Mechanics}} &
{\color{DimMemory}\textbf{Memory}} \\
\midrule
Genie 3 & -- & 0.4700 & 0.6608 & 0.6719 & 0.4711 & \textbf{0.4757} & \textbf{0.5454} & \underline{0.6073} \\
HappyOyster & -- & \textbf{0.5317} & \textbf{0.7631} & \textbf{0.7737} & \textbf{0.5235} & 0.4377 & \underline{0.5395} & \textbf{0.6309} \\
LingBot-World & 14B & 0.3235 & 0.4198 & 0.4094 & 0.2898 & 0.2875 & 0.2777 & 0.3006 \\
HY-World 1.5 & 8.3B & 0.1640 & 0.2088 & 0.2015 & 0.1400 & 0.1236 & 0.1115 & 0.1562 \\
\midrule
ABot-World-0 & 5B & \underline{0.5266} & \underline{0.7290} & \underline{0.6752} & \underline{0.5039} & \underline{0.4651} & 0.5223 & 0.5041 \\
\bottomrule
\end{tabular}
}
\end{center}
\textit{ABot-World-0} achieves strong action fidelity, trajectory following, visual quality, physical mechanics, and memory scores.

\subsubsection{Qualitative Analysis}
\label{sec:worldroambench_qualitative}

Beyond its competitive quantitative performance, \textit{ABot-World-0} demonstrates three qualitative capabilities that are critical for practical interactive world models: sustained controllability and scene coherence over extremely long rollouts, a unified control interface that generalizes across diverse scenes and controllable characters, and plausible physical responses to actions and environmental constraints.
Across the evaluated cases, \textit{ABot-World-0} remains responsive to user commands while preserving recognizable scene structure, actor and object identity, and motion continuity, without rapidly degenerating into frozen motion, repetitive textures, or severe visual drift.
These results suggest that \textit{ABot-World-0} learns persistent and stateful interactive dynamics rather than merely generating sequences of locally plausible short clips.

\begin{figure}[p]
\centering
\includegraphics[width=0.97\textwidth,keepaspectratio]{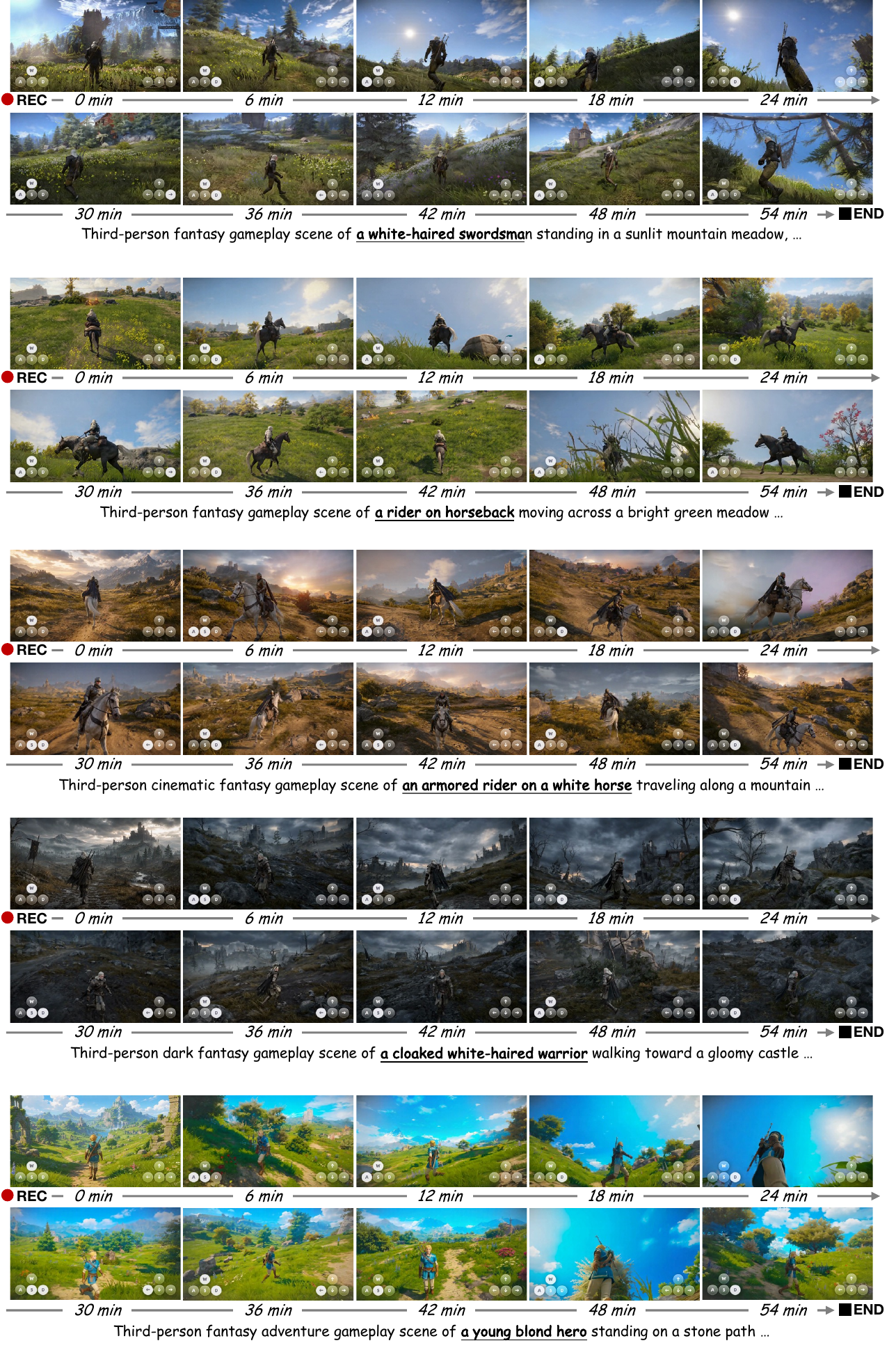}
\caption{Hour-scale long-horizon generation examples.
Each rollout is shown as a timestamped keyframe strip, demonstrating sustained controllability and scene coherence over one-hour interactive rollouts.}
\label{fig:wrb_hour_long}
\end{figure}
\begin{figure}[p]
\centering
\includegraphics[width=1\textwidth,keepaspectratio]{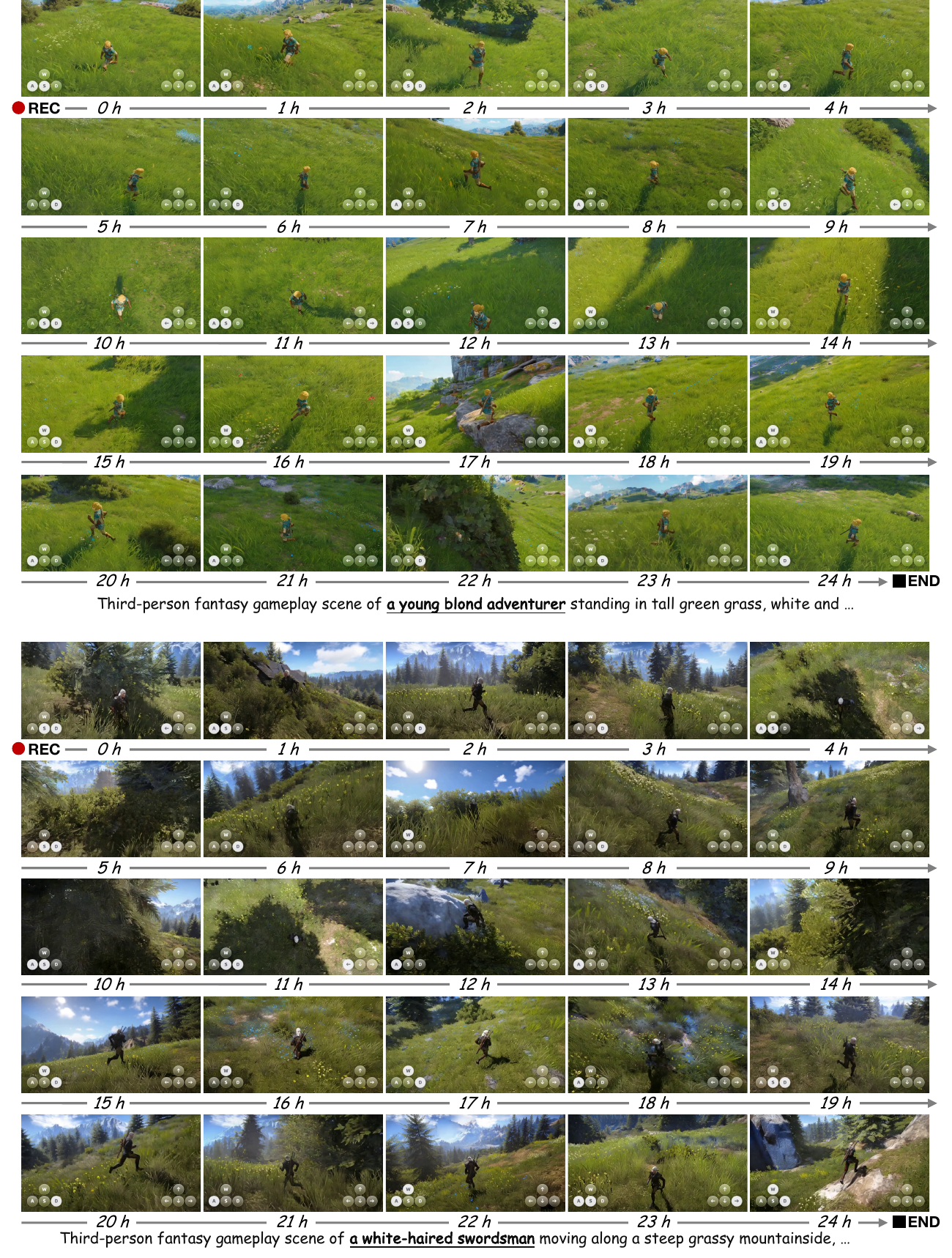}
\caption{Day-scale long-horizon generation examples, Set I.
Timestamped keyframes from two day-scale rollout examples show recognizable scene structure, viewpoint consistency, and active motion at sampled checkpoints over extended generation.}
\label{fig:wrb_day_long_1}
\end{figure}

\begin{figure}[p]
\centering
\includegraphics[width=0.97\textwidth,keepaspectratio]{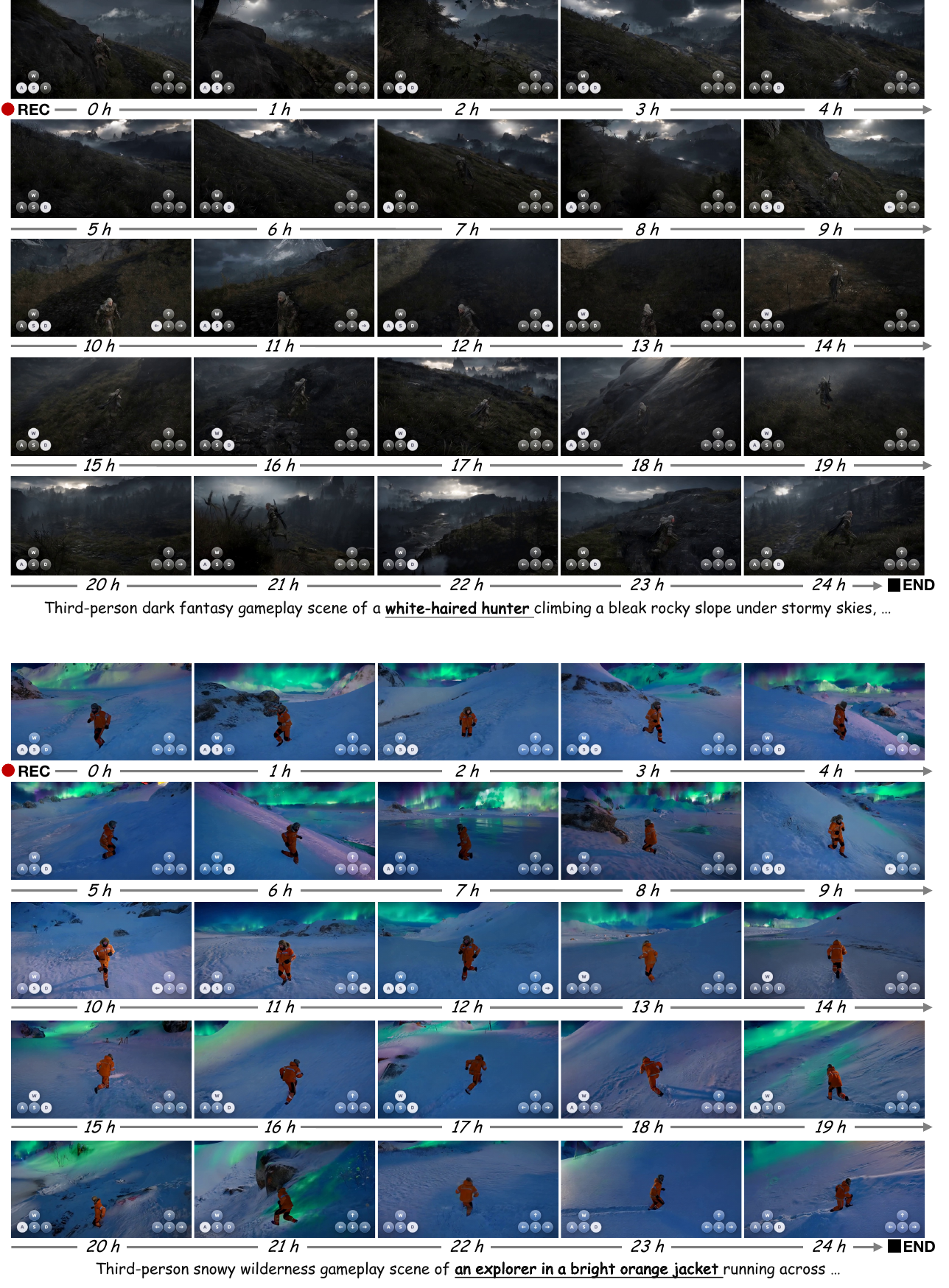}
\caption{Day-scale long-horizon generation examples, Set II.
Two additional timestamped rollout examples maintain recognizable scene structure and active motion at sampled checkpoints under substantial accumulated-error conditions.}
\label{fig:wrb_day_long_2}
\end{figure}

Figures~\ref{fig:wrb_hour_long}, \ref{fig:wrb_day_long_1}, and~\ref{fig:wrb_day_long_2} demonstrate the model's long-horizon generation capability.
The hour-scale results contain five independent rollouts visualized as timestamped keyframe strips.
Despite the extended generation horizon, \textit{ABot-World-0} preserves coherent environments, viewpoints, and controllable actors while continuously responding to the input action stream.
The day-scale stress tests further expose the model to severe accumulated-error conditions, and the sampled checkpoints retain recognizable scene structure and active motion instead of collapsing into texture noise, repetitive content, or static frames.

\begin{figure}[p]
\centering
\includegraphics[width=0.94\textwidth,keepaspectratio]{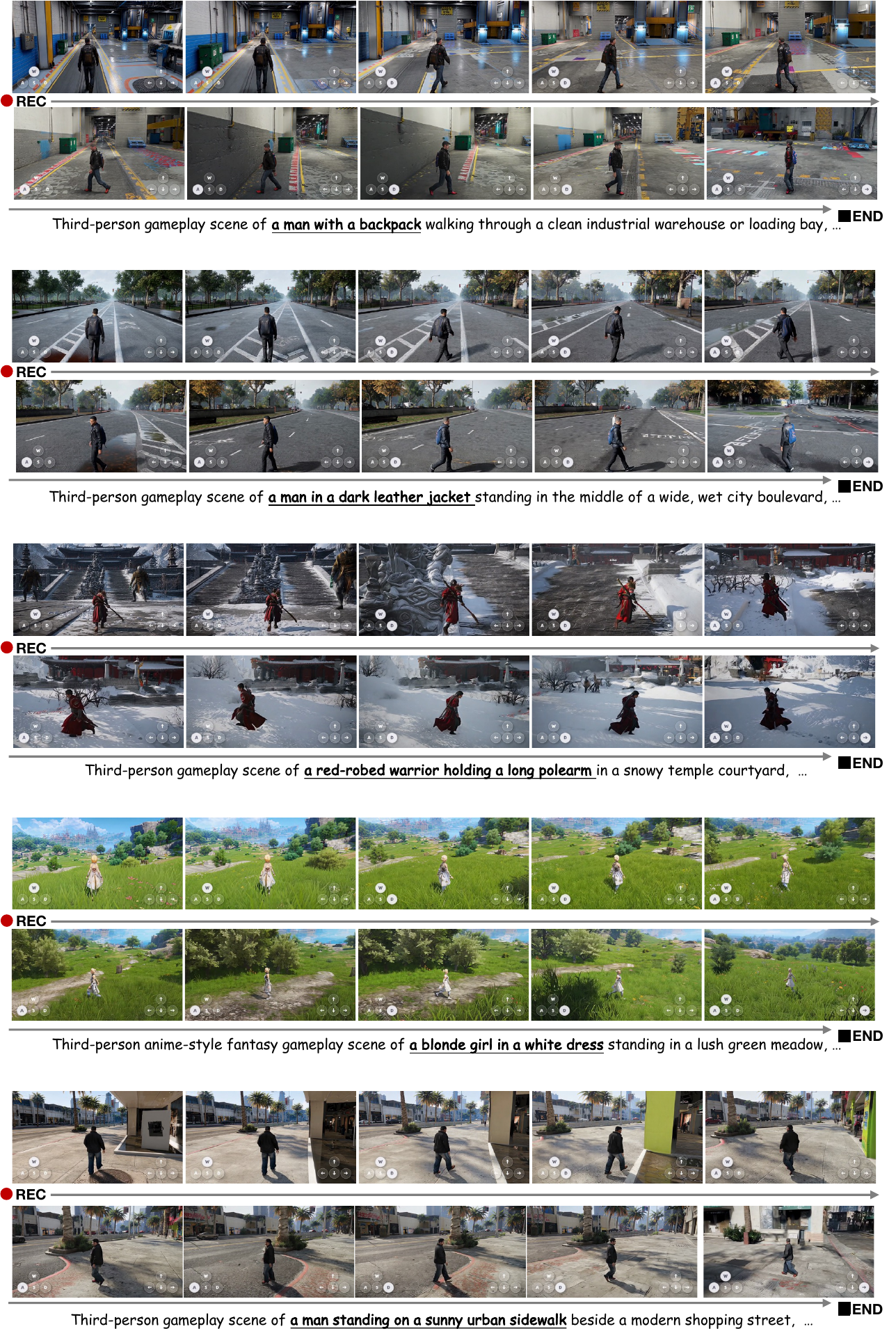}
\caption{Out-of-domain (OOD) control generalization across diverse scenes and controllable characters.
Each row presents an interactive rollout in which both the environment and the controllable subject are outside the training distribution.
Using a unified action interface, \textit{ABot-World-0} generates action-consistent character motion while preserving coherence with the surrounding scene.}
\label{fig:wrb_multi_control}
\end{figure}

Figure~\ref{fig:wrb_multi_control} evaluates out-of-domain (OOD) control generalization, where both the environment and the controllable character fall outside the training distribution.
The examples span diverse scenes and controllable characters under a unified action representation.
Across these OOD settings, the generated motion consistently follows the input action sequence while remaining coherent with the surrounding environment, demonstrating that the learned action conditioning generalizes to novel scene--character combinations.
\begin{figure}[p]
\centering
\includegraphics[width=0.94\textwidth, keepaspectratio]{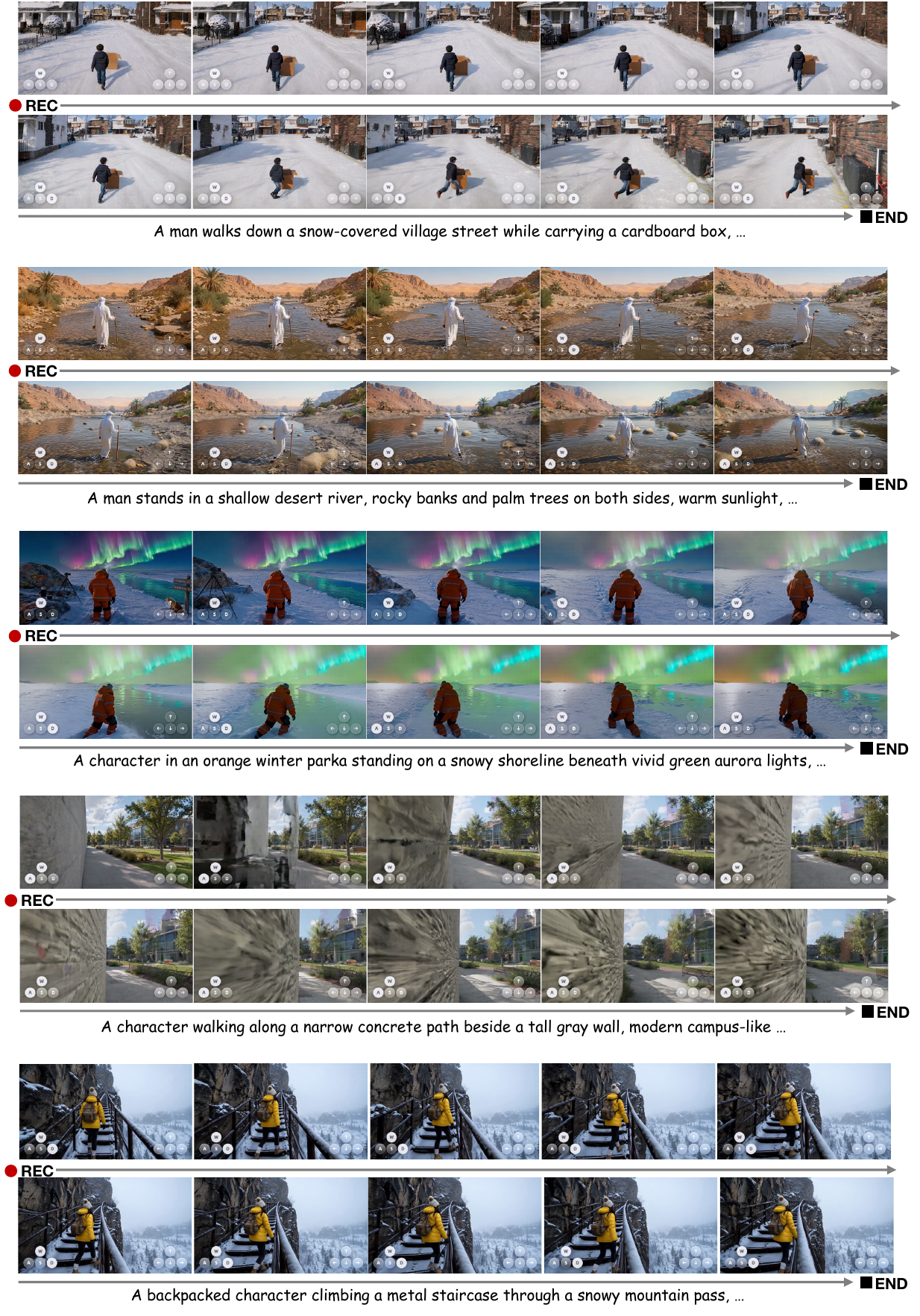}
\caption{Physical-interaction examples.
The cases include object collision, water traces, snow tracks, wall blocking, and collision with a railing without interpenetration.}
\label{fig:wrb_physics}
\end{figure}

Figure~\ref{fig:wrb_physics} highlights plausible physical consequences that emerge during interaction.
For example, a person pushes a cardboard box away, footsteps induce temporally consistent disturbances in water, a person walking through snow leaves persistent footprints, first-person motion is blocked by a wall, and a character collides with a railing without passing through it.
These cases involve collisions, contact effects, persistent environmental changes, and geometric constraints that are not specified by the action labels alone.
Although \textit{ABot-World-0} is not explicitly trained with symbolic physical rules or real-world collision annotations, it produces plausible physical responses from large-scale interactive video experience.

\clearpage
\subsection{Effect of LongForcing on Long-Horizon Rollouts}

We compare \textit{LongForcing} with a Causal-Forcing-style baseline~\cite{zhu2026causalforcing} adapted to our interactive world model under the same rollout protocol.
Both variants train on histories produced by the student's own autoregressive rollout and apply DMD in the final post-training stage.
\textit{LongForcing} differs by using extended-horizon teacher supervision during this stage rather than the shorter-horizon supervision used by the baseline.
We evaluate both variants over 60-second rollouts using the HPSv3 score~\cite{ma2025hpsv3}, high-saturation pixel ratio, perceptual blur score, and patch repeat ratio.
Higher HPSv3 is better; lower values are preferred for the other three metrics.

\begin{figure}[!htbp]
    \centering
    \captionsetup{type=figure,hypcap=false}
    \includegraphics[width=\textwidth]{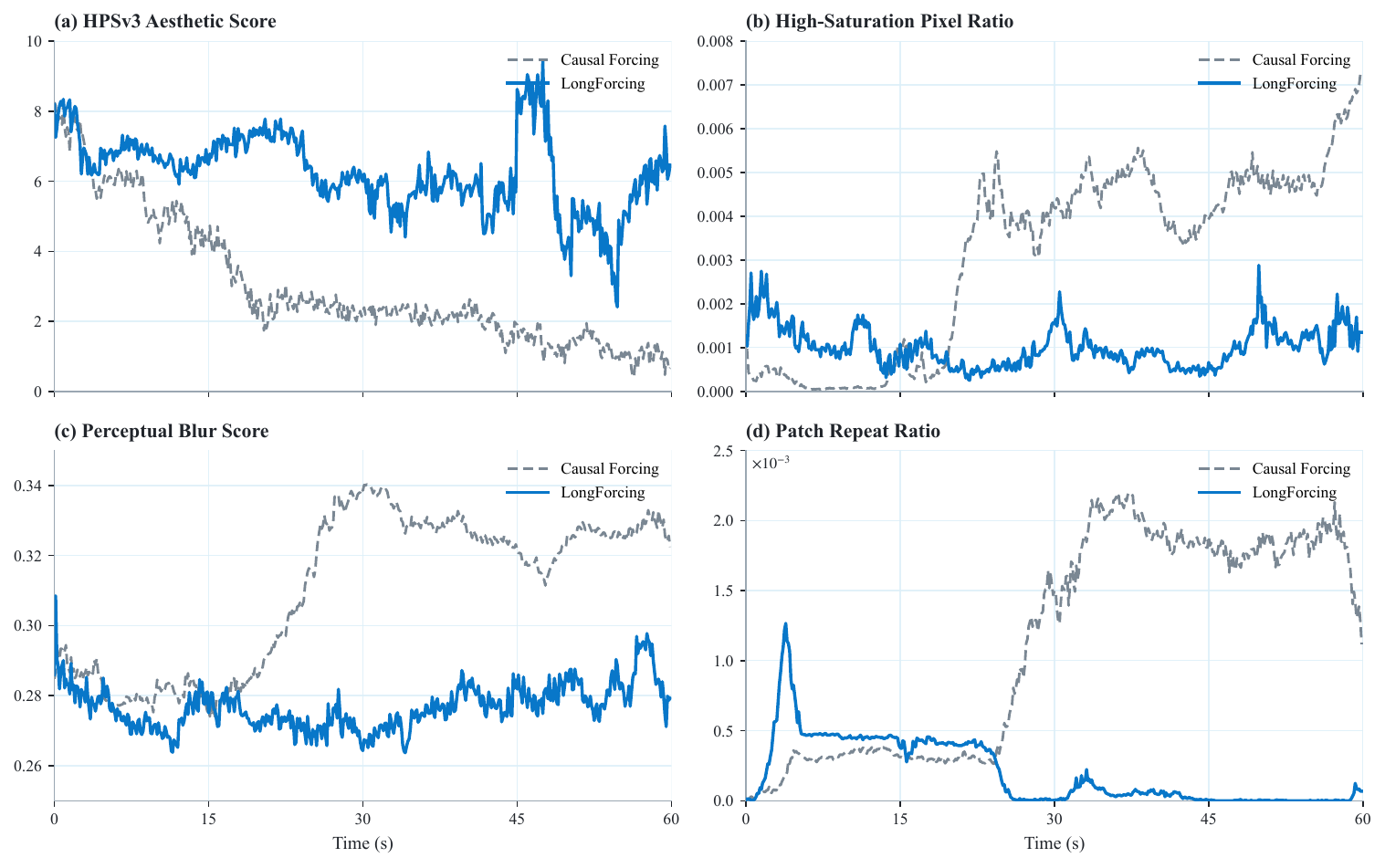}
    \caption{Framewise comparison of our Causal-Forcing-style baseline and \textit{LongForcing} over 60-second rollouts.
    Both variants use student self-rollouts and final-stage DMD; \textit{LongForcing} additionally uses extended-horizon teacher supervision.
    The curve labeled \emph{Causal Forcing} denotes the adapted Causal-Forcing-style baseline used in this report.
    Higher HPSv3 values are better, while lower values are better for high-saturation pixel ratio, perceptual blur score, and patch repeat ratio.
    Curves are averaged over the evaluation set.}
    \label{fig:longforcing_ablation}
\end{figure}

As shown in Figure~\ref{fig:longforcing_ablation}, the differences between the two variants become more pronounced during later portions of the rollout.
The Causal-Forcing-style baseline shows a progressive decrease in HPSv3 together with higher saturation, blur, and patch repetition in the second half of the rollout.
\textit{LongForcing} retains a higher HPSv3 score and lower artifact-related metrics over the same interval, indicating less visual error accumulation during autoregressive generation.
\clearpage
\section{Discussion and Future Work}
\label{sec:discussion}

\textbf{Explicit actions admit simple conditioning.}\quad
Keyboard inputs are discrete, temporally aligned with the video, and directly reflect user intent.
For these explicit control signals, we use additive injection at the patch-embedding stage to provide reliable control while preserving the visual prior of the pretrained model.
More elaborate conditioning may remain useful for ambiguous signals such as latent actions, continuous camera trajectories, or semantic instructions.

\textbf{Long-horizon drift is better understood as a distribution-shift problem.}\quad
As autoregressive generation proceeds, small errors accumulate in the visual context and gradually move the model beyond the states covered during short-horizon training.
Sink-based context stabilization and fixed reference frames can delay this process, but excessive anchoring may keep the model too close to its initial observation, restricting motion and scene evolution.
\textit{LongForcing} takes a different route: it trains on long student self-rollouts and uses an extended-horizon teacher to regularize the resulting distribution.
This allows the model to continue imagining new content while remaining stable.
The coherent sampled checkpoints in our day-scale rollouts are consistent with the objective of training-time distribution alignment as a promising route to long-horizon generation.

\textbf{Real-time interaction is more than few-step sampling.}\quad
Another practical lesson is that reducing denoising steps does not by itself produce an interactive system.
Once the DiT is distilled, VAE decoding, attention computation, memory transfer, and the memory footprint and bandwidth demands of the KV cache remain material systems costs.
Optimizing these components together enables \textit{ABot-World-0} to sustain real-time inference on a single desktop GPU.
For interactive world models, action-to-first-frame latency, sustained throughput, and memory footprint are therefore more meaningful than the number of denoising steps alone.

\textbf{Future work.}\quad
These observations point to several natural extensions.
Richer actions and semantic events may require more structured conditioning; multi-scale \textit{LongForcing} and persistent scene memory may further improve long-horizon consistency; and more efficient decoding and context management could extend high-resolution real-time generation to a broader range of consumer hardware.

\section{Conclusion}

\textit{ABot-World-0} brings action-conditioned video world modeling to real-time desktop deployment.
Trained on multi-source action-video data spanning AAA games, simulation engines, and internet videos, a single model supports both scene navigation and character control across different environments and controllable subjects.
Reference-character conditioning provides persistent appearance cues during long-horizon third-person generation.

The training pipeline starts from a bidirectional teacher that learns action-conditioned visual dynamics, and progressively converts it into an efficient causal generator through teacher forcing and ODE distillation.
\textit{LongForcing} further aligns the student's long self-rollouts with an extended-horizon teacher, extending distribution-level supervision to long-horizon student-rollout contexts encountered during repeated autoregressive generation.
The 60-second rollout evaluation, which extends beyond the training horizon, shows that \textit{LongForcing} reduces visual error accumulation relative to the Causal-Forcing-style baseline, supporting improved visual stability over the evaluated rollouts.

The optimized deployment stack combines a lightweight VAE decoder, low-bit inference, efficient attention, and memory-aware scheduling to deliver 720P streaming at up to 16 FPS with 1.2\,s action-to-first-frame latency on a single RTX 5090, while peak VRAM remains below 19.3 GiB across the optimized low-bit operating points.
Results on WorldRoamBench show competitive performance in action following, trajectory following, visual quality, physical mechanics, and memory retention.
Notably, sampled checkpoints from day-scale rollouts preserve recognizable visual quality, active dynamics, and scene coherence without observable collapse at the evaluated timestamps.

\textit{ABot-World-0} demonstrates that unified interactive control, stable long-horizon inference, and consumer-grade real-time deployment can be achieved within a single video world model.
By continuously extending the generated world beyond predefined scene boundaries, it provides an open and practical foundation for interactive content creation, game simulation, agent learning, and embodied-AI research.

\section{Contributors}

\team{Project Sponsors}
{Mu Xu and Ning Guo.}

\team{Foundation Model Team}
{Fan Jiang\leadfootnotemark, Zhaoxu Sun, Mengchao Wang, Ziyu Zhu, Chiyu Wang, Yunpeng Zhang, Wenlin Liu, Yun Wang, Xue Zheng, Rui Sun, and Junfeng Ni.}

\team{Data Team}
{Hongyu Pan\lead, Zhongxu Sun, Fei Yu, Zengye Ge, Mengmeng Du, Nianfei Fan, Mingchao Sun, Yu Liu, and Yongchang.}

\team{AI Infra Team}
{Yanqing Zhu\lead, Jiahang Wang, Ning Ying, Yuze Xuan, and Di Yang.}

\team{Benchmark Team}
{Zhicheng Liu\lead, Zhe Gao, Tingbing Xu, Jiacheng Sui, and Wenjin Yang.}

\team{Engineering Team}
{Junnan Lai\lead, Nianfei Fan, Shufeng Liu, Mengmeng Du, Yuan Liu, Zheng Zhou, Yingliang Peng, Dawei Cao, Kaifeng Sheng, Yuxiang Cai, and Fei Lu.}

\begingroup
\renewcommand{\thefootnote}{\ensuremath{\dagger}}
\footnotetext{Denotes the Tech Lead of each team.}
\endgroup

\section*{Acknowledgements}

\noindent
We sincerely thank Chengzhen Yu, Chong Sun, Chunnuo Gong, Jian Zhang, Qianwei Wang, and Yu Lei (listed alphabetically by first name) for their invaluable support, insightful feedback, and contributions throughout the development of this project.


\bibliographystyle{unsrt}
\bibliography{references}

\end{document}